\documentclass[10pt,twocolumn,letterpaper]{article}

\usepackage[pagenumbers]{cvpr} 

\usepackage[dvipsnames]{xcolor}

\newcommand{\parsection}[1]{\noindent\textbf{#1}:}
\usepackage{multirow}
\usepackage{placeins}
\usepackage{siunitx}

\usepackage{xr}
\newcommand*{\addFileDependency}[1]{
\typeout{(#1)}
\@addtofilelist{#1}
\IfFileExists{#1}{}{\typeout{No file #1.}}
}\makeatother

\usepackage[accsupp]{axessibility}

\definecolor{cvprblue}{rgb}{0.21,0.49,0.74}
\usepackage[pagebackref,breaklinks,colorlinks,citecolor=cvprblue]{hyperref}

\def\paperID{3493} 
\def\confName{CVPR}
\def\confYear{2024}

\title{NeuRAD: Neural Rendering for Autonomous Driving}

\author{Adam Tonderski$^{\dagger,1,4}$
\quad Carl Lindström$^{\dagger,1,2}$
\quad Georg Hess$^{\dagger,1,2}$
\\ William Ljungbergh$^{1,3}$
\quad Lennart Svensson$^{2}$
\quad Christoffer Petersson$^{1,2}$
\\
\normalsize$^1$Zenseact \hspace{0.8cm} $^2$Chalmers University of Technology \hspace{0.8cm} $^3$Linköping University \hspace{0.8cm} $^4$Lund University\\
{\tt\small \{firstname.lastname\}@\{zenseact.com, chalmers.se\}}
}

\def \modelname{NeuRAD}
\begin{document}

\twocolumn[{
\maketitle
\vspace{-12mm}
\begin{center}
    \captionsetup{type=figure}
    \includegraphics[width=0.95\textwidth,trim={0cm 0cm 0cm 0cm},clip]{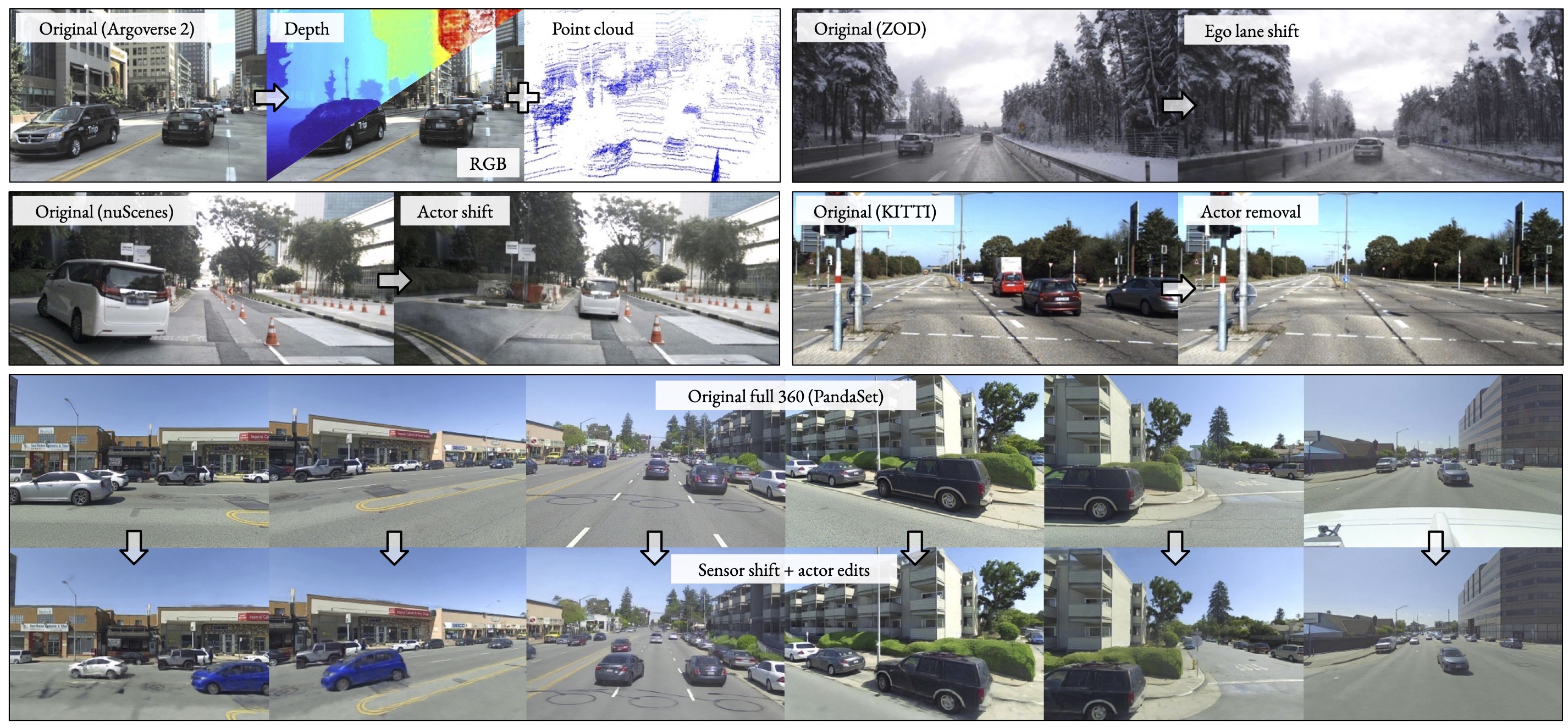}
    \vspace{-1mm}
    \captionof{figure}{
    {\modelname} is a neural rendering method tailored to dynamic automotive scenes. With it, we can alter the pose of the ego vehicle and other road users as well as freely add and/or remove actors. These capabilities make {\modelname} suitable to serve as the foundation in components such as sensor-realistic closed-loop simulators or powerful data augmentation engines.}
    \label{fig:frontpage}
\end{center}
}]

\begin{abstract}
Neural radiance fields (NeRFs) have gained popularity in the autonomous driving (AD) community. Recent methods show NeRFs' potential for closed-loop simulation, enabling testing of AD systems, and as an advanced training data augmentation technique. However, existing methods often require long training times, dense semantic supervision, or lack generalizability. This, in turn, hinders the application of NeRFs for AD at scale. In this paper, we propose {\modelname}, a robust novel view synthesis method tailored to dynamic AD data. Our method features simple network design, extensive sensor modeling for both camera and lidar -- including rolling shutter, beam divergence and ray dropping -- and is applicable to multiple datasets out of the box.  We verify its performance on five popular AD datasets, achieving state-of-the-art performance across the board. To encourage further development, we openly release the {\modelname} \href{https://github.com/georghess/NeuRAD}{source code}.
\end{abstract}

\section{Introduction}
\label{sec:intro}

In Neural Radiance Fields (NeRFs) a model is trained to learn a 3D representation from which sensor realistic data can be rendered from new viewpoints~\cite {mildenhall2021nerf}. Such techniques have been shown to be useful for a multitude of applications, such as view synthesis~\cite{barron2022mip360}, generative modeling~\cite{instructnerf2023}, or pose and shape estimation~\cite{wen2023bundlesdf}.

\def\thefootnote{$\dagger$}\footnotetext{These authors contributed equally to this work.}\def\thefootnote{\arabic{footnote}}

Autonomous Driving (AD) is a field where NeRFs may become very useful. By creating editable digital clones of traffic scenes, safety-critical scenarios can be explored in a scalable manner and without risking physical damage. For example, practitioners can investigate the behavior of the system for harsh braking on a highway or aggressive merging in city traffic. Furthermore, a NeRF-powered closed-loop simulator can be used for the targeted generation of corner-case training data.

Multiple works have applied NeRFs to automotive data~\cite{ost2021neural,kundu2022panoptic,tancik2022block,xie2023snerf,wu2023mars,unisim,turki2023suds}. Neural Scene Graphs~\cite{ost2021neural} extend the original NeRF model~\cite{mildenhall2021nerf} to dynamic automotive sequences by dividing the scene into static background and a set of rigid dynamic actors with known location and extent, learning separate NeRFs for each. This enables editing the trajectories of both the ego-vehicle and all actors in the scene. The approach can be further improved by including semantic segmentation~\cite{kundu2022panoptic} or by using anti-aliased positional embeddings~\cite{xie2023snerf}. The latter enables NeRFs to reason about scale~\cite{barron2022mip360} which is essential for large-scale scenes. However, common for all these approaches is that they require many hours of training, limiting their applicability for scalable closed-loop simulation or data augmentation.

More recent works~\cite{unisim,wu2023mars} rely on Instant NGP's (iNGP)~\cite{muller2022instant} learnable hash grids for embedding positional information, drastically reducing training and inference time. Further, these methods generate realistic renderings in their respective settings, namely front-facing camera with 360$^\circ$ lidar. However, their performance in $360^\circ$ multicamera settings, which is common in many AD datsets~\cite{caesar2020nuscenes,Argoverse2}, is either unexplored~\cite{wu2023mars} or is reported by the authors to be suboptimal~\cite{unisim}. Furthermore, both methods deploy simple lidar models and cannot model ray drop, a phenomenon important for closing the real-to-sim gap~\cite{manivasagam2023towards}. Lastly, using the iNGP positional embedding without anti-aliasing techniques limits performance, especially for larger scenes~\cite{Barron_2023_ICCV}.  

In this paper, we present {\modelname}, an editable novel view synthesis (NVS) method, designed to handle large-scale automotive scenes and to work well with multiple datasets off the shelf. We find that modeling sensor characteristics, such as rolling shutter, lidar ray dropping, and beam divergence, is essential for sensor-realistic renderings. Further, our model features a simple network architecture, where static and dynamic elements are discerned only by their positional embeddings, making it a natural extension of recent methods to AD data. We verify {\modelname}'s generalizability and achieve state-of-the-art performance across five automotive datasets, with no dataset-specific tuning.

Our contributions are as follows. \textbf{(1)} Our method is the first to combine lidar sensor modeling with the ability to handle 360$^\circ$ camera rigs in a unified way, extending the applicability of NeRF-based methods for dynamic AD data. \textbf{(2)} We propose using a single network to model dynamic scenes, where dynamics and statics are separated only by their positional embeddings. \textbf{(3)} We propose simple, yet effective methods for modeling multiple key sensor characteristics such as rolling shutter, beam divergence, and ray dropping, and highlight their effect on performance. \textbf{(4)} Extensive evaluation using five popular AD datasets shows that our method is state-of-the-art across the board. 

\section{Related work}
\label{sec:related_work}

\parsection{NeRFs} Neural radiance fields~\cite{mildenhall2021nerf} is a novel view synthesis method in which a neural network learns an implicit 3D representation from which new images can be rendered. Multiple works~\cite{muller2022instant,chen2022tensorf,fridovich2022plenoxels,kerbl20233d} address the long training time of the original formulation. Notably, Instant-NGP (iNGP)~\cite{muller2022instant} uses a multiresolution, learnable hash grid to encode positional information rather than NeRFs frequency-based encoding scheme.
A different line of work \cite{barron2021mip,barron2022mip360,Barron_2023_ICCV,hu2023tri} focuses on reducing aliasing effects by embedding pixel frustums instead of extent-free points, where Zip-NeRF~\cite{Barron_2023_ICCV} combines the anti-aliasing properties of mip-NeRF 360~\cite{barron2022mip360} with the fast hash grid embedding of iNGP~\cite{muller2022instant} by using multisampling and downweighting.
Although these works were designed for static scenes and cannot be applied to dynamic sequences, we draw inspiration from Zip-NeRF's anti-aliasing techniques to better model large scenes.

\parsection{NeRFs for automotive data} Accurately simulating data for AD systems is a promising avenue for efficient testing and verification of self-driving vehicles. While game-engine-based methods~\cite{dosovitskiy2017carla,shah2018airsim} have made a lot of progress, they struggle with scalable asset creation, real-to-sim gap, and diversity. NeRFs' sensor-realistic renderings offer an attractive alternative, and consequently, multiple works have studied how to apply neural rendering techniques to automotive data. NSG~\cite{ost2021neural}, Panoptic Neural Fields (PNF)~\cite{kundu2022panoptic} and Panoptic NeRF~\cite{fu2022panoptic} all model the background and every actor as multi-layer perceptrons (MLPs), but struggle with large-scale scenes due to the MLPs limited expressiveness. S-NeRF~\cite{xie2023snerf} extends mip-NeRF 360 to automotive data similar to NSG by modeling each actor with a separate MLP, but requires day-long training, making it impractical for downstream applications. Block-NeRF~\cite{tancik2022block} and SUDS~\cite{turki2023suds} both focus on city-scale reconstruction. While handling impressive scale, Block-NeRF filters out dynamic objects and only models static backgrounds, and SUDS uses a single network for dynamic actors, removing the possibility of altering actor behavior. 

\parsection{NeRFs for closed-loop simulation} Among existing work, two methods~\cite{wu2023mars,unisim} are the most similar to ours. MARS~\cite{wu2023mars} proposes a modular design where practitioners can mix and match existing NeRF-based methods for rendering dynamic actors and the static background. Similar to our work, the implementation is based on Nerfstudio~\cite{tancik2023nerfstudio} to promote open-source collaboration. Unlike our work, MARS does not natively support lidar point clouds but relies on dense depth maps from either depth completion or mono-depth networks, limiting the ease of application to any dataset. Further, while MARS' semantic segmentation supervision is optional, performance deteriorates when this supervision is not available, especially on real-world data. 

UniSim~\cite{unisim} is a neural sensor simulator, showcasing realistic renderings for PandaSet's~\cite{pandaset} front camera and 360$^\circ$ lidar. The method applies separate hash grid features~\cite{muller2022instant} for modeling the sky, the static background, and each dynamic actor, and uses NSG-style~\cite{ost2021neural} transformations for handling dynamics. For efficiency, the static background is only sampled near lidar points. Further, UniSim renders features from the neural field, rather than RGB, and uses a convolutional neural network (CNN) for upsampling the features and producing the final image. This allows them to reduce the number of sampled rays per image significantly. While efficient, multiple approximations lead to poor performance outside their evaluation protocol. In addition, the lidar occupancy has a limited vertical field of view and fails to capture tall, nearby structures which often becomes evident when using cameras with alternative mounting positions or wider lenses, \eg, nuScenes~\cite{caesar2020nuscenes}, Argoverse2~\cite{Argoverse2} or Zenseact Open Dataset (ZOD)~\cite{alibeigi2023zenseact}. In contrast, our method unifies static and sky modeling and relies on proposal sampling~\cite{Barron_2023_ICCV} for modeling occupancy anywhere. Further, UniSim's upsampling CNN introduces severe aliasing and model inconsistencies, as camera rays must describe entire RGB patches whereas lidar rays are thin laser beams. In this work, we introduce a novel anti-aliasing strategy that improves performance, with minimal impact on runtime.

\section{Method}
\label{sec:method}
\begin{figure*}[tb]
    \centering
    \includegraphics[width=0.9\linewidth]{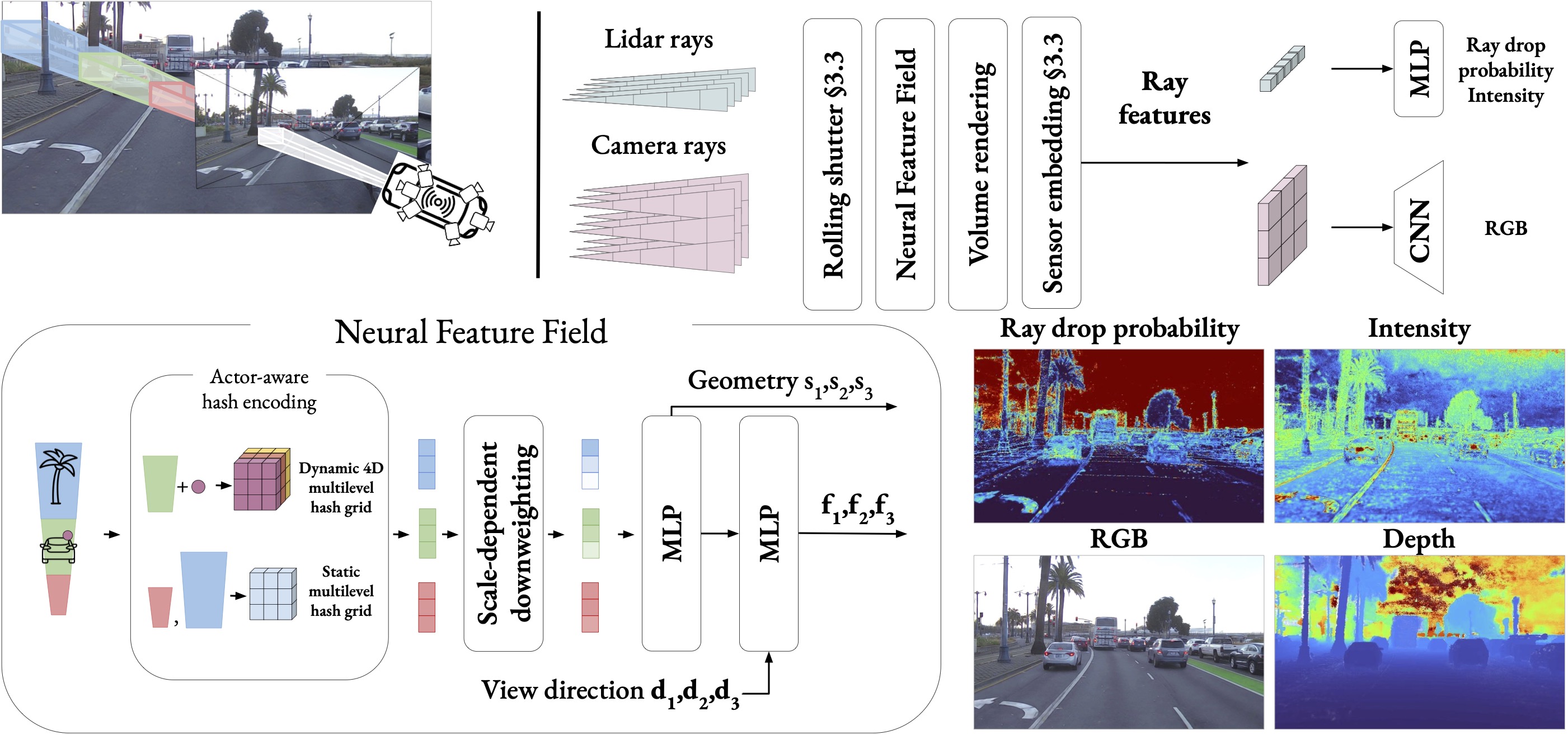}
    \caption{Overview of our approach. We learn a joint neural feature field for the statics and dynamics of an automotive scene, where the two are discerned only by our actor-aware hash encoding. Points that fall inside actor bounding boxes are transformed to actor-local coordinates and, together with actor index, used to query the 4D hash grid. We decode the volume rendered ray-level features to RGB values using an upsampling CNN, and to ray drop probability and intensity using MLPs.}
    \vspace{-5mm}
    \label{fig:method-overview}
\end{figure*}

Our goal is to learn a representation from which we can generate realistic sensor data where we can change either the pose of the ego vehicle platform, the actors, or both. We assume access to data collected by a moving platform, consisting of posed camera images and lidar point clouds, as well as estimates of the size and pose of any moving actors. To be practically useful, our method needs to perform well in terms of reconstruction error on any major automotive dataset, while keeping training and inference times to a minimum. To this end, we propose {\modelname}, an editable, open source, and performant neural rendering approach; see \cref{fig:method-overview} for an overview.

In the following, we first describe the underlying scene representation and sensor modeling. Next, we cover the internals of our neural field and the decomposition of sequences into static background and dynamic actors. We then present the unique challenges and opportunities of applying neural rendering to AD data and how we address them. Last, we discuss learning strategies.

\subsection{Scene representation and sensor modeling}

\parsection{Neural scene rendering}
Building on the recent advancements in novel view synthesis~\cite{unisim, Barron_2023_ICCV}, we model the world with a neural feature field (NFF), a generalization of NeRFs \cite{mildenhall2021nerf} and similar methods~\cite{mescheder2019occupancy}. Given a position $\mathbf{x}$, and a view direction $\mathbf{d}$, an NFF outputs an implicit geometry $s$ and a feature vector $\mathbf{f}$~\cite{unisim}. The NFF, akin to a NeRF, is utilized for volumetric rendering. However, it accumulates implicit geometry and features rather than density and color~\cite{mildenhall2021nerf}.

To extract features for a ray $\mathbf{r}(\tau) = \mathbf{o} + \tau \mathbf{d}$, originating from the sensor center $\mathbf{o}$ and extending in direction $\mathbf{d}$, we sample $N_\mathbf{r}$ points along the ray in 3D space. The feature descriptors of these samples are aggregated using traditional alpha compositing:
\begin{equation} \label{eq:feature_aggregation}
    \vspace{-2mm}
    \mathbf{f}(\mathbf{r}) = \sum^{N_\mathbf{r}}_{i=1} w_i \mathbf{f}_i, \quad w_i = \alpha_i \prod^{i-1}_{j=1}(1 - \alpha_j).
\end{equation}
Here, $\alpha_i$ represents the opacity at the point $\mathbf{x}_i=\mathbf{o} +\tau_i \mathbf{d}$, and $w_i$ the opacity times the accumulated transmittance along the ray up to $\mathbf{x}_i$. Inspired by its success in recovering high-quality geometry~\cite{oechsle2021unisurf,li2023neuralangelo}, we represent the implicit geometry using a signed distance function (SDF) and approximate the opacity as
$\alpha_i =1/({1 + e^{\beta s_i}})$,
where $s_i$ is the SDF value at $\mathbf{x}_i$ and $\beta$ is a learnable parameter. While more accurate SDF formulations~\cite{wang2021neus,wang2023neus2} can provide better performance, they require gradient calculations for each 3D point, negatively impacting the runtime.

\parsection{Camera modeling} To render an image, we volume render a set of camera rays, generating a feature map $\mathcal{F}\in\mathbb{R}^{H_f \times W_f \times N_f}$. As in \cite{unisim}, we then rely on a CNN to render the final image $\mathcal{I}\in\mathbb{R}^{H_\mathcal{I} \times W_\mathcal{I} \times 3}$. In practice, the feature map has a lower resolution $H_f \times W_f$ than the image $H_I \times W_I$, and we use the CNN for upsampling. This allows us to drastically reduce the number of queried rays.

\parsection{Lidar modeling} Lidar sensors allow self-driving vehicles to measure the depth and the reflectivity (intensity) of a discrete set of points. They do so by emitting laser beam pulses and measuring the time of flight to determine distance and returning power for reflectivity. To capture these properties, we model the transmitted pulses from a posed lidar sensor as a set of rays and use volume rendering similar to \eqref{eq:feature_aggregation}. For a lidar point, we shoot a ray $\mathbf{r}(\tau) = \mathbf{o} + \tau \mathbf{d}$, where $\mathbf{o}$ is the origin of the lidar and $\mathbf{d}$ is the normalized direction of the beam. We then find the expected depth $D_l$ of a ray as $\mathbb{E}[{D}_l(\mathbf{r})] = \sum^{N_\mathbf{r}}_{i=1} w_i \tau_i$.
For predicting intensity, we volume render the ray feature following \eqref{eq:feature_aggregation} and pass the feature through a small MLP.

In contrast to previous works incorporating lidar measurements \cite{unisim,rematas2022urban}, we also include rays for laser beams which did not return any points. This phenomenon, known as ray dropping, occurs if the return power has too low amplitude, and is important to model for reducing the sim-to-real gap~\cite{manivasagam2023towards}. Typically, such rays travel far without hitting a surface, or hit surfaces from which the beam bounces off into empty space, \eg, mirrors, glass, or wet road surfaces. Modeling these effects is important for sensor-realistic simulations, but as noted in \cite{huang2023nfl}, are hard to capture fully physics-based because they depend on (often undisclosed) details of the low-level sensor detection logic. Therefore, we opt to learn ray dropping from data. Similar to the intensity, we use the rendered ray feature from \eqref{eq:feature_aggregation} and pass it through a small MLP to predict the ray drop probability $p_d(\mathbf{r})$. Note that unlike \cite{huang2023nfl}, we do not model second returns from lidar beams, as this information is not present in the five datasets considered here.

\subsection{Extending Neural Feature Fields}

In this section, we delve into the specifics of our volumetric scene representation. We begin by extending the Neural Feature Field (NFF) definition to be a learned function \( (s, \mathbf{f}) = \text{NFF}(\mathbf{x}, t, \mathbf{d}) \), where \( \mathbf{x} \in \mathbb{R}^3 \) are the spatial coordinates, \( t \in \mathbb{R} \) represents time, and \( \mathbf{d} \in \mathbb{R}^3 \) indicates the view direction. Importantly, this definition introduces time as an input, which is essential for modeling the dynamic aspects of the scene.

\parsection{Architecture}
Our NFF architecture adheres to well-established best practices in the NeRF literature \cite{Barron_2023_ICCV, muller2022instant}. Given a position \( \mathbf{x} \) and time \( t \) we query our \textit{actor-aware hash encoding}. This encoding then feeds into a small Multilayer Perceptron (MLP), which computes the signed distance \( s \) and an intermediate feature \( \mathbf{g} \in \mathbb{R}^{N_g} \). The view direction \( \mathbf{d} \) is encoded using spherical harmonics \cite{muller2022instant}, allowing the model to capture reflections and other view-dependent effects. Finally, the direction encoding and \( \mathbf{g} \) are jointly processed through a second MLP, augmented with a skip connection from \( \mathbf{g} \), producing the feature $\mathbf{f}$.

\parsection{Scene composition}
Similar to previous works~\cite{unisim,kundu2022panoptic,ost2021neural,xie2023snerf}, we decompose the world into two parts, the static background and a set of rigid dynamic actors, each defined by a 3D bounding box and a set of SO(3) poses. This serves a dual purpose: it simplifies the learning process, and it allows a degree of editability, where actors can be moved after training to generate novel scenarios. Unlike previous methods which utilize separate NFFs for different scene elements, we employ a single, unified NFF, where all networks are shared, and the differentiation between static and dynamic components is transparently handled by our actor-aware hash encoding. The encoding strategy is straightforward: depending on whether a given sample (\( \mathbf{x}, t \)) lies inside an actor bounding box, we encode it using one of two functions.

\parsection{Unbounded static scene}
We represent the static scene with a multiresolution hash grid~\cite{muller2022instant}, as this has been proven to be a highly expressive and efficient representation. However, to map our unbounded scenes onto a grid, we employ the contraction approach proposed in MipNerf-360~\cite{barron2022mip360}. This allows us to accurately represent both nearby road elements and far-away clouds, with a single hash grid. In contrast, prior automotive approaches utilize a dedicated NFF to capture the sky and other far-away regions~\cite{unisim}.

\parsection{Rigid dynamic actors}
When a sample (\( \mathbf{x}, t \)) falls within the bounding box of an actor, its spatial coordinates \( \mathbf{x} \) and view directions \( \mathbf{d} \) are transformed to the actor's coordinate frame at the given time $t$. This allows us to ignore the time aspect after that, and sample features from a time-independent multiresolution hash grid, just like for the static scene. Naively, we would need to separately sample multiple different hash grids, one for each actor. However, we instead utilize a single 4D hash grid, where the fourth dimension corresponds to the actor index. This novel approach allows us to sample all actor features in parallel, achieving significant speedups while matching the performance of using separate hash grids.

\subsection{Automotive data modeling}
\label{sec:auto_data_modeling}
\parsection{Multiscale scenes}
One of the biggest challenges in applying neural rendering to automotive data is handling the multiple levels of detail present in this data. As vehicles cover large distances, many surfaces are visible both from afar and close up. Applying iNGP's~\cite{muller2022instant} or NeRF's position embedding naively in these multiscale settings results in aliasing artifacts as they lack a sense at which scale a certain point is observed~\cite{barron2021mip}. To address this, many approaches model rays as conical frustums, the extent of which is determined longitudinally by the size of the bin and radially by the pixel area in conjunction with distance to the sensor ~\cite{barron2021mip,barron2022mip360,hu2023tri}. Zip-NeRF~\cite{Barron_2023_ICCV}, which is currently the only anti-aliasing approach for iNGP's hash grids, combines two techniques for modeling frustums: multisampling and downweighting. In multisampling, the positional embeddings of multiple locations in the frustum are averaged, capturing both longitudinal and radial extent. For downweighting, each sample is modeled as an isotropic Gaussian, and grid features are weighted proportional to the fraction between their cell size and the Gaussian variance, effectively suppressing finer resolutions. While the combined techniques significantly increase performance, the multisampling also drastically increases run-time.

Here, we aim to incorporate scale information with minimal run-time impact. Inspired by Zip-NeRF, we propose an intuitive downweighting scheme where we downweight hash grid features based on their size relative to the frustum. Rather than using Gaussians, we model each ray $\mathbf{r}(\tau)=\mathbf{o}+\tau\mathbf{d}$ as a pyramid with cross-sectional area $A(\tau)=\dot{r}_h \dot{r}_v \tau^2$, where $\dot{r}_h, \dot{r}_v$ are horizontal and vertical beam divergence based on the image patch size or the beam divergence of the lidar beam. Then, for a frustum defined by the interval $[\tau_i, \tau_{i+1})$, where $A_i$ and $A_{i+1}$ are the cross-sectional areas at the end-points $\tau_i$ and $\tau_{i+1}$, we calculate its volume as
\begin{equation}
    V_i = \frac{\tau_{i+1}-\tau_i}{3}\left(A_i+\sqrt{A_iA_{i+1}}+A_{i+1}\right),
\end{equation}
and retrieve its positional embedding $\mathbf{e}_i$ at the 3D point $\mathbf{x}_i=\mathbf{o}+\frac{\tau_i+\tau_{i+1}}{2}\mathbf{d}$. Finally, for a hash grid at level $l$ with resolution $n_l$ we weight the position embedding $\mathbf{e}_{i,l}$ with $\omega_{i,l}=\texttt{min}(1,(\frac{1}{n_l V_i^{1/3}}))$, \ie, the fraction between the cell size and the frustum size.

\parsection{Efficient Sampling} Another difficulty with rendering large-scale scenes is the need for an efficient sampling strategy. In a single image, we might want to render detailed text on a nearby traffic sign while also capturing parallax effects between skyscrapers several kilometers away. Uniformly sampling the ray to achieve both of these goals would require thousands of samples per ray which is computationally infeasible. Previous works have relied heavily on lidar data for pruning samples~\cite{unisim}, and as a result struggle to render outside the lidar's field-of-view.

Instead, we draw samples along rays according to a power function \cite{Barron_2023_ICCV}, such that the space between samples increases with the distance from the ray origin. Even so, we find it impossible to fulfill all relevant conditions without prohibitively increasing the number of samples. Therefore, we also employ two rounds of proposal sampling~\cite{mildenhall2021nerf}, where a lightweight version of our NFF is queried to generate a weight distribution along the ray. Then, a new set of samples are drawn according to these weights. After two rounds of this procedure, we are left with a refined set of samples that focus on the relevant locations along the ray and that we can use to query our full-size NFF. To supervise the proposal networks, we adopt an anti-aliased online distillation method~\cite{Barron_2023_ICCV} and further use the lidar for supervision, see $\mathcal{L}^\text{d}$ and $\mathcal{L}^\text{w}$ introduced in \cref{sec:losses}.

\parsection{Modeling rolling shutter}
In standard NeRF-based formulations, each image is assumed to be captured from a single origin $\mathbf{o}$. However, many camera sensors have rolling shutters, \ie, pixel rows are captured sequentially. Thus, the camera sensor can move between the capturing of the first row and that of the last row, breaking the single origin assumption. Although not an issue for synthetic data~\cite{mildenhall2019local} or data captured with slow-moving handheld cameras, the rolling shutter becomes evident with captures from fast-moving vehicles, especially for side-cameras. The same effect is also present in lidars, where each scan is typically collected over \SI{0.1}{\second}, which corresponds to several meters when traveling at highway speeds. Even for ego-motion compensated point clouds, these differences can lead to detrimental line-of-sight errors where 3D points translate to rays that cut through other geometries. To mitigate these effects, we model the rolling shutters by assigning individual times to each ray and adjusting their origin according to the estimated motion. As the rolling shutter affects all dynamic elements of the scene, we linearly interpolate actor poses to each individual ray time. See \cref{sec:rolling_shutter} for details.

\begin{figure*}
    \centering
    \includegraphics[width=\textwidth]{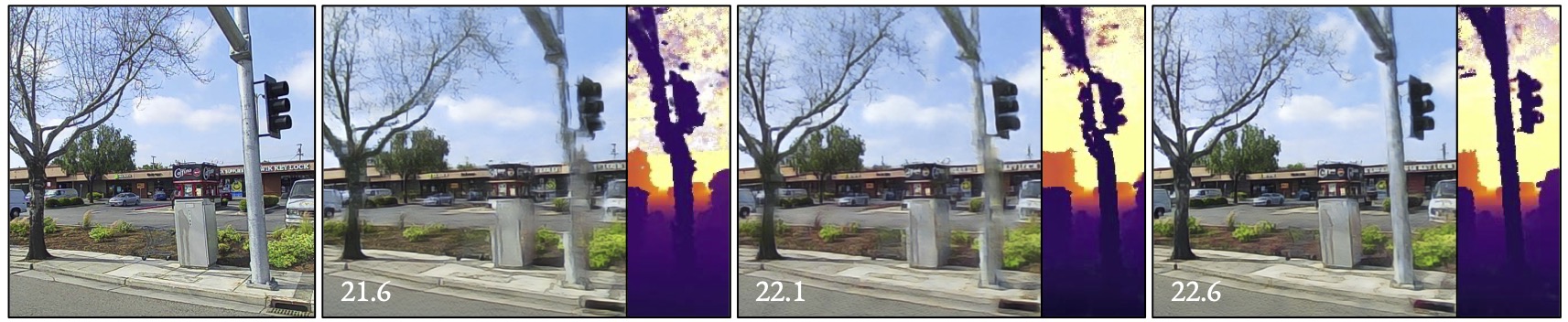}
    \begin{subfigure}[t]{0.18\textwidth}
        \vspace{-4.5mm}
        \subcaption{original}
        \label{fig:sub1}
    \end{subfigure}%
    \begin{subfigure}[t]{0.27\textwidth}
        \vspace{-4.5mm}
        \subcaption{no modeling}
        \label{fig:sub2}
    \end{subfigure}%
    \begin{subfigure}[t]{0.27\textwidth}
        \vspace{-4.5mm}
        \subcaption{modeling lidar only}
        \label{fig:sub3}
    \end{subfigure}%
    \begin{subfigure}[t]{0.27\textwidth}
        \vspace{-4.5mm}
        \subcaption{modeling lidar + camera}
        \label{fig:sub4}
    \end{subfigure}
    \vspace{-3mm}
    \caption{Impact of modeling rolling shutter in a high-speed scenario (with inset PSNR). (a) original side-camera image. Omitting the rolling shutter entirely (b) results in extremely blurry renderings and unrealistic geometry, especially for the pole. Modeling the lidar rolling shutter (c) improves the quality, but it is only when both sensors are modeled correctly (d) that we get realistic renderings.}
    \label{fig:qualitative results}
    \vspace{-3mm}
\end{figure*}

\parsection{Differing camera settings} Another problem when modeling autonomous driving sequences is that images come from different cameras with potentially different capture parameters, such as exposure. Here we draw inspiration from research on ``NeRFs in the wild''~\cite{martin2021nerf}, where an appearance embedding is learned for each image, and passed to the second MLP together with $\mathbf{g}$. However, as we know which image comes from which sensor, we instead learn a single embeddings per sensor, minimizing the potential for overfitting, and allowing us to use these \textit{sensor embeddings} when generating novel views. As we render features rather than color, we apply these embeddings after the volume rendering, significantly reducing computational overhead.

\parsection{Noisy actor poses}
Our model relies on estimates of poses for dynamic actors, either in the form of annotations or as tracking output. To account for imperfections, we include the actor poses as learnable parameters in the model, and optimize them jointly. The poses are parameterized as a translation $\mathbf{t}\in\mathbb{R}^3$ and a rotation for which we use a 6D-representation~\cite{zhou2019continuity}.

\subsection{Losses}
\label{sec:losses}
We optimize all model components jointly and use both camera and lidar observations as supervision $\mathcal{L} = \mathcal{L}^\text{image} + \mathcal{L}^\text{lidar}$.
In the following, we discuss the different optimization objectives in more detail.

\parsection{Image losses}
The image loss is computed patch-wise and summed over $N_p$ patches and consists of a reconstruction term $\mathcal{L}^\text{rgb}$ and a perceptual term $\mathcal{L}^\text{vgg}$:
\begin{equation}
\vspace{-2mm}
    \mathcal{L}^\text{image} = \frac{1}{N_p} \sum_{i=1}^{N_p} \lambda^\text{rgb} \mathcal{L}^\text{rgb}_i + \lambda^\text{vgg} \mathcal{L}^\text{vgg}_i .
    \vspace{-1mm}
\end{equation}
The reconstruction loss is the squared error between predicted and true pixel values. The perceptual loss is the distance between VGG features for real and predicted patches \cite{wang2018pix2pixHD}. $\lambda^\text{rgb}$ and $\lambda^\text{vgg}$ are weighting hyperparameters.

\parsection{Lidar losses}
We incorporate the strong geometric prior given by the lidar by adding a depth loss for lidar rays and employing weight decay to penalize density in empty space. Further, to be able to simulate a more realistic lidar we also include objectives for the predicted intensity and the predicted ray drop probability:
\begin{multline}
\vspace{-2.2mm}
    \mathcal{L}^\text{lidar} = \frac{1}{N} \sum_{i=1}^{N}(
    \lambda^\text{d} \mathcal{L}^\text{d}_i
    + \lambda^\text{int} \mathcal{L}^\text{int}_i
    + \lambda^{p_d} \mathcal{L}^{p_d}_i
    + \lambda^\text{w} \mathcal{L}^\text{w}_i ), 
    \vspace{-2.2mm}
\end{multline}
where $\lambda^{\text{d}}$, $\lambda^{\text{int}}$, $\lambda^{p_d}$, and $\lambda^{\text{w}}$ are hyperparameters. The depth loss $\mathcal{L}^\text{d}_i$ and the intensity loss $\mathcal{L}^\text{int}_i$ are the squared error between the prediction and the observation. For dropped rays, we penalize estimates only below the specified sensor range, and do not supervise intensity. For the ray drop probability loss, $\mathcal{L}^{p_d}_i$, we use a binary cross entropy loss. The weight decay is applied for all samples outside of a distance $\epsilon$ of the lidar observation:
\vspace{-1.5mm}
\begin{equation}
    \mathcal{L}^\text{w}_i = \sum_{\tau_{i,j} > \epsilon} \left \| w_{ij}  \right \|_2,
    \vspace{-1.5mm}
\end{equation}
where $\tau_{i,j}$ is the distance from sample $\mathbf{x}_{ij}$ to the lidar observation for ray $i$. For dropped rays, weight decay is applied up until the specified sensor range. Noteably, we omit the commonly used eikonal loss, as it provided minimal benefits at a high computational cost.

\subsection{Implementation details}
{\modelname} is implemented in the collaborative, open-source project Nerfstudio~\cite{tancik2023nerfstudio}. We hope that our developed supporting structures such as data loaders and native lidar support will encourage further research into this area. We train our main method (NeuRAD) for 20,000 iterations using the Adam~\cite{adam} optimizer. Using a single Nvidia A100, training takes about 1 hour. To showcase the scalability of our approach, we also design a larger model with longer training (NeuRAD-2x). See \cref{sec:training_details} for further details.

\section{Experiments}
\label{sec:experiments}
To verify the robustness of our model, we evaluate its performance on several popular AD datasets: nuScenes~\cite{caesar2020nuscenes}, PandaSet~\cite{pandaset}, Argoverse 2~\cite{Argoverse2}, KITTI~\cite{geiger2013vision}, and ZOD~\cite{alibeigi2023zenseact}. To prove the robustness of our method we use the same model and hyperparameters on all datasets. We investigate novel view synthesis performance both for hold-out validation images and for sensor poses without any ground truth. Furthermore, we ablate important model components. More results, including a study on the real2sim gap as well as failure cases can be found in \cref{sec:simulation_gap} and \cref{sec:more_results}.

\begin{table}[tb]
    \centering
    \caption{Image novel view synthesis performance comparison to state-of-the-art methods across five datasets. *our reimplementation. $^\dagger$baselines from \cite{unisim,wu2023mars,xie2023snerf}. $^\mathsection$partial results due to training instability. \textbf{Bold}/\underline{underline} for best/second-best.}
    \vspace{-2mm}
    \resizebox{0.85\linewidth}{!}{
    \begin{tabular}{lcccc}
    \toprule
    &     &  PSNR $\uparrow$ & SSIM $\uparrow$ & LPIPS $\downarrow$ \\ \midrule
    \multirow{5}{*}{\rotatebox[origin=c]{90}{\shortstack[c]{\small Panda\\FC}}}& Instant-NGP$^\dagger$ \cite{muller2022instant,unisim}    & 24.03             & 0.708             & 0.451 \\
    & UniSim \cite{unisim} & 25.63 & 0.745 & 0.288 \\
    &  UniSim* & 25.44 & 0.732 & 0.228 \\
    & {\modelname} (ours) & \underline{26.58} & \underline{0.778} & \underline{0.190} \\
    & {\modelname}-2x (ours) & \textbf{26.84} & \textbf{0.801} & \textbf{0.148} \\
    \midrule
    \multirow{3}{*}{\rotatebox[origin=c]{90}{\shortstack[c]{\small Panda\\360}}}& UniSim* & 23.50 & 0.692 & 0.330 \\
    & {\modelname} (ours) & \underline{25.97} & \underline{0.758} & \underline{0.242} \\
    & {\modelname}-2x (ours) & \textbf{26.47} & \textbf{0.779} & \textbf{0.196} \\
    \midrule
    \multirow{4}{*}{\rotatebox[origin=c]{90}{\shortstack[c]{\small nuScenes}}}& Mip360$^\dagger$ \cite{barron2022mip360,xie2023snerf} & 24.37 & 0.795 & 0.240 \\
    & S-NeRF \cite{xie2023snerf} & 26.21 & \textbf{0.831} & 0.228 \\
    & {\modelname} (ours) & \underline{26.99} & 0.815 & \underline{0.225} \\
    & {\modelname}-2x (ours) & \textbf{27.13} & \underline{0.820} & \textbf{0.205} \\
    \midrule
    \multirow{4}{*}{\rotatebox[origin=c]{90}{\shortstack[c]{\small KITTI\\MOT}}}&  SUDS$^\dagger$ \cite{turki2023suds,wu2023mars} & 23.12 & \underline{0.821} & 0.135 \\
    & MARS \cite{wu2023mars} & 24.00 & 0.801 & 0.164 \\
    & {\modelname} (ours) & \underline{27.00} & 0.795 & \underline{0.082} \\
    & {\modelname}-2x (ours) & \textbf{27.91} & \textbf{0.822} & \textbf{0.066} \\
    \midrule
    \multirow{3}{*}{\rotatebox[origin=c]{90}{\shortstack[c]{\small Argo2}}}& UniSim* & 23.22$^\mathsection$ & 0.661$^\mathsection$ & 0.412$^\mathsection$ \\
    & {\modelname} (ours) & \underline{26.22} & \underline{0.717} & \underline{0.315} \\
    & {\modelname}-2x (ours) & \textbf{27.73} & \textbf{0.756} & \textbf{0.233} \\
    \midrule
    \multirow{3}{*}{\rotatebox[origin=c]{90}{\shortstack[c]{\small ZOD}}}& UniSim* & 27.97 & 0.777 & 0.239 \\
    & {\modelname} (ours) & \underline{29.49} & \underline{0.809} & \underline{0.226} \\
    & {\modelname}-2x (ours) & \textbf{30.59} & \textbf{0.857} & \textbf{0.210} \\
    \bottomrule
    \end{tabular}
    }
    \vspace{-5mm}
    \label{tab:sota_nvs_image}
\end{table}

\subsection{Datasets and baselines}
Below, we introduce the datasets used for evaluation. The selected datasets cover various sensors, and the included sequences contain different seasons, lighting conditions, and driving conditions. Existing works typically use one or two datasets for evaluation and build models around assumptions about available supervision, limiting their applicability to new settings. Therefore, for each dataset, we compare our model to SoTA methods that have previously adopted said dataset, and follow their respective evaluation protocols. Similar to our method, UniSim~\cite{unisim} imposes few supervision assumptions, and we, therefore, reimplement the method (denoted Unisim$^*$) and use it as a baseline for datasets where no prior work exists. See \cref{sec:unisim_implementation_details} for reimplementation details and \cref{sec:evaluation_details} for further evaluation details.

\parsection{PandaSet}
We compare our method to UniSim~\cite{unisim} and an iNGP version with lidar depth supervision provided by UniSim. We use every other frame for training and the remaining ones for testing, and evaluate on the same 10 scenes as UniSim. We study two settings: one with lidar and front-facing camera (Panda FC) for direct comparison with the results reported in \cite{unisim}, and one with lidar and all six cameras capturing the full 360$^\circ$ field-of-view around the vehicle (Panda 360). We also evaluate UniSim on the full 360$^\circ$ setting using our reimplementation.

\parsection{nuScenes}
We compare our method to S-NeRF~\cite{xie2023snerf} and Mip-NeRF 360~\cite{barron2022mip360}. We follow S-NeRF's protocol, \ie, select 40 consecutive samples halfway into the sequences and use every fourth for evaluation while every other among the remaining ones is used for training. We test on the same four sequences as S-NeRF, using the same sensor setup.

\parsection{KITTI}
For KITTI~\cite{geiger2013vision}, we compare our method to MARS~\cite{wu2023mars}. We use MARS 50\% evaluation protocol, \ie, evaluating on every second image from the right camera and using the left and right camera and lidar from remaining time instances for training.

\parsection{Argo 2 \& ZOD}
To verify the robustness of our method, we study two additional datasets, Argoverse 2~\cite{Argoverse2} and ZOD~\cite{alibeigi2023zenseact}. Due to the lack of prior work supporting dynamic actors on these datasets, we compare {\modelname} to our UniSim implementation. For each dataset, we train on every other frame, test on the remaining frames, and evaluate on ten sequences. As ZOD does not have any sequence annotations, we use a 3D-object detector and an off-the-shelf tracker to generate pseudo-annotations for the sequences.

\begin{figure}[tb]
    \centering
    \includegraphics[width=1.0\linewidth]{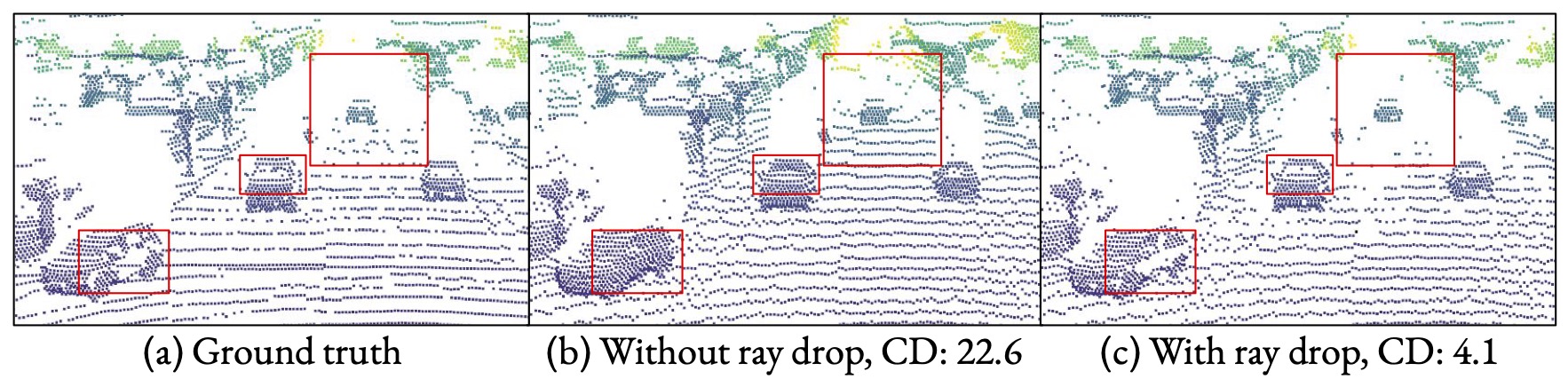}
    \vspace{-7mm}
    \caption{Visualization of ray drop effects for lidar simulation. Highlighted parts show areas where ray dropping effects are important to consider in order to simulate realistic point clouds. CD denotes Chamfer distance normalized by num. GT points.}
    \label{fig:lidar_ray_drop}
    \vspace{-2mm}
\end{figure}

\subsection{Novel view synthesis}

\parsection{Camera}
We report the standard NVS metrics PSNR, SSIM~\cite{wang2004image} and LPIPS~\cite{zhang2018unreasonable}, for all datasets and baselines in \cref{tab:sota_nvs_image}. {\modelname} achieves SoTA performance across all datasets. On PandaSet, we improve upon previous work across all metrics, for both FC and 360. On nuScenes, {\modelname} matches the performance of S-NeRF while training much faster (1 hour compared to 17 hours). {\modelname} also outperforms previous SoTA on KITTI with a large margin in terms of PSNR and LPIPS. Finally, {\modelname} also achieves strong performance on Argoverse 2 and ZOD.

\begin{table}[t]
    \centering
    \caption{Lidar novel view synthesis performance comparison to state-of-the-art methods. Depth is median L2 error [\si{\m}]. Intensity is RMSE. Drop acc. denotes ray drop accuracy. Chamfer denotes chamfer distance, normalized with num. ground truth points [\si{\m}].}
    \vspace{-2mm}
    \resizebox{0.99\linewidth}{!}{
    \begin{tabular}{lccccc}
    \toprule
    &     &  Depth $\downarrow$ & Intensity $\downarrow$ & Drop acc. $\uparrow$ & Chamfer $\downarrow$ \\ \midrule
    \multirow{3}{*}{\rotatebox[origin=c]{90}{\shortstack[c]{\small Panda\\FC}}} & UniSim    & 0.10          & 0.065 & -              & -            \\
    & UniSim*                                                                               & 0.07          & 0.085          & 91.0           & 11.2         \\
    & {\modelname} (ours)                                                                           & \textbf{0.01} & \textbf{0.062}          & \textbf{96.2}           & \textbf{1.6}          \\ \midrule
    \multirow{2}{*}{\rotatebox[origin=c]{90}{\shortstack[c]{\small Panda\\360}}} & UniSim*  & 0.07          & 0.087          & 91.9           & 10.3         \\
    & {\modelname} (ours)                                                                           & \textbf{0.01} & \textbf{0.061}          & \textbf{96.1}  & \textbf{1.9} \\
    \bottomrule
    \end{tabular}
    }
    \vspace{-4mm}
    \label{tab:sota_nvs_lidar}
\end{table}

\parsection{Lidar}
We measure the realism of our lidar simulation in terms of L2 median depth error, RMSE intensity error and ray drop accuracy. We complement the depth error with the Chamfer distance as it enables us to evaluate performance on dropped rays as well. We compare only to UniSim, evaluated on PandaSet, as no other baseline simulates point clouds. UniSim has no notion of ray dropping, hence we assume rays to be dropped past the reported lidar range. We see in \cref{tab:sota_nvs_lidar} that {\modelname} decreases the depth error by an order of magnitude compared to UniSim in the front-camera setting. Our method generalizes well to the 360$^\circ$ setting, where similar results are reported. Furthermore, we show that {\modelname} is capable of simulating realistic point clouds, thanks to its high ray drop accuracy and low Chamfer distance. \cref{fig:lidar_ray_drop} further shows the importance of modeling ray drop effects for lidar simulation. As noted in the figure, lidar beams that hit the road far away tend to disperse and not return. Similar effects occur for transparent surfaces, such as the car window illustrated in the figure, where the lidar beams shoot right through. Modeling these effects can increase the realism of simulated point clouds.

\subsection{Novel scenario generation}
In order for our method to be useful in practice, it must not only perform well when interpolating between views, but also when exploring new views, as examplified in \cref{fig:frontpage}. To that end, we investigate {\modelname}'s capability to generate images from poses that are significantly different from those encountered during training. We adapt UniSim's protocol on PandaSet, \ie, translating the ego vehicle sensors laterally two or three meters to simulate a lane shift, and extend the protocol to include one meter vertical shift, simulating other mounting positions. We further investigate ``actor shift'', and rotate ($\pm 0.5$ radians) or translate ($\pm 2$ meters laterally) dynamic actors in the scene to simulate different actor behaviors. As no ground truth images exist, we report FID~\cite{fid}, with ``no shift'' for reference. The results in \cref{tab:extrapolation_image} show that {\modelname} is able to generalize to new viewpoints and learns meaningful actor representations. We also include results where we optimize the camera poses following~\cite{wang2021nerf}, as this further increases sharpness.

\begin{table}[t]
    \centering
    \caption{FID scores when shifting pose of ego vehicle or actors.}
    \vspace{-2mm}
    \resizebox{0.99\linewidth}{!}{
    \begin{tabular}{lccccccc}
    \toprule
    \multirow{2}{*}{} & \multirow{2}{*}{} & \multicolumn{4}{c}{Ego shift} & \multicolumn{2}{c}{Actor shift} \\
    \cmidrule(lr){3-6} \cmidrule(lr){7-8}
         & & No shift & Lane 2\si{\m} & Lane 3\si{\m} & Vert. 1\si{\m} & Rot. & Trans. \\ \midrule
         \multirow{3}{*}{\rotatebox[origin=c]{90}{\shortstack[c]{\small Panda\\FC}}}  & UniSim      & -             & 74.7   & 97.5 & - & - & - \\
                                                                               & UniSim*     & 41.7          & 79.6   & 102.0 & 89.3 & 65.5 & 59.6 \\
                                                                               & {\modelname}        & \textbf{25.0} & \textbf{72.3} & \textbf{93.9} & \textbf{76.3} & \textbf{64.3} & \textbf{49.1} \\\midrule
         \multirow{3}{*}{\rotatebox[origin=c]{90}{\shortstack[c]{\small Panda\\360}}} & UniSim*     & 88.3          & 115.5         & 128.0         & 126.7         & 95.9          & 93.0 \\
                                                                               & {\modelname}        & 45.5          & 84.0          & 98.8          & 91.3          & 58.8          & 55.4 \\
                                                                               & {\modelname} w/ opt & \textbf{43.0} & \textbf{81.0} & \textbf{95.3} & \textbf{88.8} & \textbf{56.7} & \textbf{53.0} \\
         \bottomrule
    \end{tabular}
    }
    \label{tab:extrapolation_image}
    \vspace{-2mm}
\end{table}

\subsection{Ablations}
We validate the effectiveness of some key components in \cref{tab:ablations}. To avoid biases toward any specific dataset, we report averaged metrics from sequences from all five datasets considered in this work. We select 4 diverse sequences from each dataset, see details in \cref{sec:evaluation_details}. Our full model corresponds to the model used in all prior experiments and strikes a good balance between run-time and performance. We see that the CNN decoder (a) significantly increases both quality and speed, by requiring significantly fewer rays and allowing for interaction between rays. Accurate sensor modeling is also very important, as each of our contributions in that area provide complementary performance boost: considering rolling shutter (b) or lidar rays that did not return (e), modeling each ray as a frustum (c) and per-sensor appearance embeddings (d). We also demonstrate that replacing individual actor hash grids with a single 4D hash grid (f) has no detrimental impact on quality, while significantly increasing training speed. Finally, we replace our SDF with a NeRF-like density formulation (g). The performance is overall almost identical and shows that our model can be configured to either of these field representations depending on the need. If we desire to extract surfaces from our model, we can use an SDF, but if our scenes are dominated by fog, transparent surfaces, or other effects where an SDF breaks down, we can fall back to a density formulation. Interestingly, our ablations only show a modest impact of considering rolling shutter. However, upon closer inspection of the qualitative results, see \cref{fig:qualitative results}, it is apparent that both the renderings and underlying geometry break down without considering this effect.

\begin{table}[t]
    \centering
    \caption{Ablations when \textit{removing} core parts of our model. We report NVS performance for images and lidars, scene generation, and training megapixels per second (MP/s). Results are averaged over 20 sequences, evenly split across all five datasets.}
    \addtolength{\tabcolsep}{-2pt}
    \vspace{-2mm}
    \resizebox{0.99\linewidth}{!}{
    \begin{tabular}{clcccccc}
    \toprule
        &                   & PSNR $\uparrow$  & LPIPS $\downarrow$ & SSIM $\uparrow$ & Depth $\downarrow$ & Scen. gen. $\downarrow$ & MP/s $\uparrow$  \\
       & Full model       & 27.26 & 0.213 & 0.786 & 0.030 & 75.5  & 1.9  \\ \midrule
    a) & CNN decoder      & 25.29 & 0.329 & 0.720 & 0.107 & 127.9 & 0.2  \\
    b) & Rolling shutter  & 26.77 & 0.246 & 0.763 & 0.060 & 80.6  & 1.9  \\
    c) & Downweighting    & 26.12 & 0.283 & 0.741 & 0.146 & 100.6 & 2.0  \\
    d) & Appearance emb.  & 25.50 & 0.270 & 0.744 & 0.080 & 102.6 & 1.9  \\
    e) & Missing points   & 25.36 & 0.361 & 0.685 & 0.050 & 106.3 & 1.8  \\
    f) & 4D actor grid    & 27.22 & 0.217 & 0.779 & 0.030 & 76.5  & 1.5  \\
    g) & SDF              & 27.37 & 0.211 & 0.790 & 0.029 & 75.5  & 1.9  \\
    \bottomrule
    \end{tabular}
    }
    \addtolength{\tabcolsep}{2pt}
    \label{tab:ablations}
    \vspace{-2mm}
\end{table}

\section{Conclusions}
\label{sec:conclusions}

In this paper, we have proposed {\modelname}, a neural simulator tailored specifically for dynamic autonomous driving (AD) data. The model jointly handles lidar and camera data in 360$^\circ$ and decomposes the world into its static and dynamic elements, allowing the creation of sensor-realistic editable clones of real world driving scenarios. {\modelname} incorporates novel modeling of various sensor phenomena including beam divergence, ray dropping, and rolling shutters, all increasing the quality during novel view synthesis. We demonstrate {\modelname}'s efficacy and robustness by obtaining state-of-the-art performance on five publicly AD datasets, using a single set of hyperparameters. Lastly, we publicly release our source-code to foster more research into NeRFs for AD.

\parsection{Limitations} {\modelname} assumes actors to be rigid and does not support any deformations. Further, many modeling assumptions are invalid for harsh weather like heavy rain or snow. We hope to address these limitations in future work. 

\parsection{Acknowledegments}
We thank Maryam Fatemi for valuable feedback. This work was partially supported by the Wallenberg AI, Autonomous Systems and Software Program (WASP) funded by the Knut and Alice Wallenberg Foundation. Computational resources were provided by NAISS at \href{https://www.nsc.liu.se/}{NSC Berzelius}, partially funded by the Swedish Research Council, grant agreement no. 2022-06725.
\FloatBarrier
{
    \small
    \bibliographystyle{ieeenat_fullname}
    \bibliography{main}
}

\clearpage
\maketitlesupplementary
\appendix

In the supplementary material, we provide implementation details for our method and baselines, evaluation details, and additional results. In \cref{sec:training_details}, we describe our network architecture more closely and provide hyperparameter values. In \cref{sec:evaluation_details}, we provide details on the experimental setting. Then, in \cref{sec:unisim_implementation_details}, we provide details on our baseline implementation. We closely describe the process of inferring lidar rays that did not return in \cref{sec:app_raydrop}. Next, we cover additional details of our proposed rolling shutter modeling in \cref{sec:rolling_shutter}. Last, in \cref{sec:more_results}, we showcase additional results and highlight some limitations of our method.

\section{Implementation details}
\label{sec:training_details}

Here we describe our model and training in more detail.

\parsection{Learning} We train all parts of our model jointly for 20,000 iterations, using the Adam optimizer. In each iteration, we randomly sample 16,384 lidar rays, and 40,960 camera rays, the latter corresponding to 40 ($32\times32$) patches. For most parameters, we use a learning rate of 0.01, with a short warmup of 500 steps. For the actor trajectory optimization and the CNN decoder, we adopt a longer warmup of 2500 steps, and a lower learning rate of 0.001. If enabled, camera optimization uses a learning rate of 0.0001, also with a warmup of 2500. We use learning rate schedules that decay the rate by an order of magnitude over the course of the training.

\parsection{Networks} As we primarily compare our method with UniSim~\cite{unisim}, we follow their network design to a large degree. Our first (geo) MLP has one hidden layer, our second (feature) MLP has two hidden layers, and the lidar decoder also has two hidden layers. For details on the CNN decoder, we refer to \cref{sec:unisim:model}. All networks use a hidden dimension of 32, which is also the dimensionality of the intermediate NFF features.

\parsection{Hashgrids} We use the efficient hashgrid implementation from tiny-cuda-nn \cite{tiny-cuda-nn}, with two separate hashgrids for the static scene and the dynamic actors. We use a much larger hash table for the static world, as actors only occupy a small portion of the scene, see \cref{tab:hyperparameters}.

\parsection{Proposal Sampling} First, we draw uniform samples according to the power function $\mathcal{P}(0.1x, -1.0)$~\cite{Barron_2023_ICCV}, where we have adjusted the parameters to better match our automotive scenes. Next, we perform two rounds of proposal sampling, represented by two separate density fields. Both fields use our actor-aware hash encoding, but with smaller hash tables and a feature dimension of one in the hash tables. Instead of an MLP, we decode density with a single linear layer. The proposal fields are supervised with the anti-aliased online distillation approach proposed for ZipNeRF~\cite{Barron_2023_ICCV}. Additionally, we supervise lidar rays directly with $\mathcal{L}^\text{d}$ and $\mathcal{L}^\text{w}$.

\parsection{NeuRAD-2x} We upscale NeuRAD in a straightforward manner -- by doubling the size of all hash tables, thereby approximately doubling the model’s parameter count. As this model is primarily intended for long sequences and large scenes, we also double the resolution of each level of the static hashgrid. To accommodate the expanded model complexity, we extend the training to 50,000 iterations and adjust the warm-up periods correspondingly. All other hyperparameters remain the same. We find that while further scaling offers benefits in some cases, it leads to diminishing returns in others.

\begin{table}[t]
    \centering
    \caption{Hyperparameters for NeuRAD.}
    \begin{tabular}{lll}
                                                                                        & Hyperparameter                & Value            \\ \midrule
        \multirow{8}{*}{\rotatebox[origin=c]{90}{\shortstack[c]{Neural feature field}}} & RGB upsampling factor         & 3                \\
                                                                                        & proposal samples              & 128, 64          \\
                                                                                        & SDF $\beta$                   & 20.0 (learnable) \\
                                                                                        & power function $\lambda$      & -1.0             \\
                                                                                        & power function scale          & 0.1              \\
                                                                                        & appearance embedding dim      & 16               \\
                                                                                        & hidden dim (all networks)     & 32               \\
                                                                                        & NFF feature dim               & 32               \\
        \midrule
        \multirow{8}{*}{\rotatebox[origin=c]{90}{\shortstack[c]{Hashgrids}}}            & hashgrid features per level   & 4                \\
                                                                                        & actor hashgrid levels         & 4                \\
                                                                                        & actor hashgrid size           & $2^{15}$         \\
                                                                                        & static hashgrid levels        & 8                \\
                                                                                        & static hashgrid size          & $2^{22}$         \\
                                                                                        & proposal features per level   & 1                \\
                                                                                        & proposal static hashgrid size & $2^{20}$         \\
                                                                                        & proposal actor hashgrid size  & $2^{15}$         \\
        \midrule
        \multirow{8}{*}{\rotatebox[origin=c]{90}{\shortstack[c]{Loss weights}}}         & $\lambda^\text{rgb}$          & 5.0              \\
                                                                                        & $\lambda^\text{vgg}$          & 5e-2             \\
                                                                                        & $\lambda^\text{int}$          & 1e-1             \\
                                                                                        & $\lambda^\text{d}$            & 1e-2             \\
                                                                                        & $\lambda^\text{w}$            & 1e-2             \\
                                                                                        & $\lambda^{p_d}$               & 1e-2             \\
                                                                                        & proposal $\lambda^\text{d}$   & 1e-3             \\
                                                                                        & proposal $\lambda^\text{w}$   & 1e-3             \\
                                                                                        & interlevel loss multiplier    & 1e-3             \\
        \midrule
        \multirow{4}{*}{\rotatebox[origin=c]{90}{\shortstack[c]{Learning                                                                   \\rates}}} & actor trajectory lr & 1e-3 \\
                                                                                        & cnn lr                        & 1e-3             \\
                                                                                        & camera optimization lr        & 1e-4             \\
                                                                                        & remaining parameters lr       & 1e-2             \\ \midrule
    \end{tabular}
    \label{tab:hyperparameters}
\end{table}

\section{Evaluation details}
\label{sec:evaluation_details}

Here, we describe in detail the evaluation protocol of each SoTA method that we have compared NeuRAD to.

\parsection{Pandaset (UniSim)}
UniSim uses a simple evaluation protocol, where the entire sequence is used, with every other frame selected for evaluation and the remaining half of the frames for training. The authors report numbers for the front camera and the 360$^\circ$ lidar on the following sequences: \texttt{001, 011, 016, 028, 053, 063, 084, 106, 123, 158}. We call this protocol Panda FC, and additionally report Panda 360 results, with all 6 cameras (and the 360$^\circ$ lidar). For the backward-facing camera, we crop away 250 pixels from the bottom of the image, as this mainly shows the trunk of the ego vehicle.

\parsection{nuScenes (S-NeRF)}
S-Nerf uses four sequences for evaluation: \texttt{0164, 0209, 0359, 0916}. The first 20 samples from each sequence are discarded, and the next 40 consecutive samples are considered for training and evaluation. The remaining samples are also discarded. Out of the selected samples, every fourth is used for evaluation and the rest are used for training. We train and evaluate on all 6 cameras.

\parsection{KITTI (MARS)}
MARS reports NVS quality on a single sequence, \texttt{0006}, on frames 5-260. We adopt their 50\%-protocol, where half of the frames are used for training, and 25\% for evaluation. Following their implementation, we adopt a repeating pattern where two consecutive frames are used for training, one is discarded, and the fourth is used for evaluation.

\parsection{Argoverse 2 \& ZOD}
Here, we use a simple evaluation protocol that is analogous to that used for PandaSet. We select 10 diverse sequences for each dataset, and use each sequence in its entirety, alternating frames for training and evaluation. For Argoverse, we use all surround cameras and both lidars on the following sequences: \texttt{
    05fa5048-f355-3274-b565-c0ddc547b315,
    0b86f508-5df9-4a46-bc59-5b9536dbde9f,
    185d3943-dd15-397a-8b2e-69cd86628fb7,
    25e5c600-36fe-3245-9cc0-40ef91620c22,
    27be7d34-ecb4-377b-8477-ccfd7cf4d0bc,
    280269f9-6111-311d-b351-ce9f63f88c81,
    2f2321d2-7912-3567-a789-25e46a145bda,
    3bffdcff-c3a7-38b6-a0f2-64196d130958,
    44adf4c4-6064-362f-94d3-323ed42cfda9,
    5589de60-1727-3e3f-9423-33437fc5da4b}.
For ZOD, we use the front-facing camera and all three lidars on the following sequences: \texttt{000784, 000005, 000030, 000221, 000231, 000387, 001186, 000657, 000581, 000619, 000546, 000244, 000811}.
As ZOD does not provide sequence annotations, we use a lidar-based object detector and create tracklets using ImmortalTracker~\cite{immortal}.

\parsection{Ablation Dataset}
We perform all ablations on 20 sequences, four from each dataset considered above. We use sequences \texttt{001, 011, 063, 106} for PandaSet, \texttt{0164, 0209, 0359, 0916} for nuScenes, \texttt{0006, 0010, 0000, 0002} for KITTI, \texttt{000030, 000221, 000657, 000005} for ZOD, and \texttt{
    280269f9-6111-311d-b351-ce9f63f88c81,
    185d3943-dd15-397a-8b2e-69cd86628fb7,
    05fa5048-f355-3274-b565-c0ddc547b315,
    0b86f508-5df9-4a46-bc59-5b9536dbde9f
} for Argoverse 2. Here, we no longer adopt the dataset-specific evaluation protocols corresponding to each SoTA method. Instead, we evaluate on the full sequences, on all available sensors, alternating frames for training and evaluation. The exception is nuScenes, where we find the provided poses to be too poor to train on the full sequences. If we optimize poses during training, we get qualitatively good results, and strong FID scores, but poor reconstruction scores due to misalignment between the learned poses and the evaluation poses, see \cref{sec:more_results} for a more detailed exposition. Therefore, we re-use S-NeRF's shortened evaluation protocol, where this problem is mostly avoided, and leave the problem of proper evaluation on nuScenes for future work.

\section{UniSim implementation details}
\label{sec:unisim_implementation_details}
UniSim~\cite{unisim} is a neural closed-loop sensor simulator. It features realistic renderings and imposes few assumptions about available supervision, \ie, it only requires camera images, lidar point clouds, sensor poses, and 3D bounding boxes with tracklets for dynamic actors. These characteristics make UniSim a suitable baseline, as it is easy to apply to new autonomous driving datasets. However, the code is closed-source and there are no unofficial implementations either. Therefore, we opt to reimplement UniSim, and as our model, we do so in Nerfstudio~\cite{tancik2023nerfstudio}. As the UniSim main article does not specify many model specifics, we rely on the supplementary material available through IEEE Xplore\footnote{\url{https://ieeexplore.ieee.org/document/10204923/media}}. Nonetheless, some details remain undisclosed, and we have tuned these hyperparameters to match the reported performance on the 10 selected PandaSet~\cite{pandaset} sequences. We describe the design choices and known differences below.

\subsection{Data processing}
\parsection{Occupancy grid dilation} UniSim uses uniform sampling to generate queries for its neural feature field. Inside dynamic actors' bounding boxes, the step size is \SI{5}{\centi\meter} and inside the static field, the step size is \SI{20}{\centi\meter}. To remove samples far from any surfaces and avoid unnecessary processing, UniSim deploys an occupancy grid. The grid, of cell size \SI{0.5}{\meter}, is initialized using accumulated lidar point clouds where the points inside the dynamic actors have been removed. A grid cell is marked occupied if at least one lidar point falls inside of it. Further, the occupancy grid is dilated to account for point cloud sparseness. We set the dilation factor to two. We find the performance to be insensitive to the selection of dilation factor, where larger values mainly increase the number of processed samples.

\parsection{Sky sampling} UniSim uses 16 samples for the sky field for each ray. We sample these linearly in disparity (one over distance to the sensor origin) between the end of the static field and \SI{3}{\kilo\meter} away.

\parsection{Sample merging} Each ray can generate a different number of sample points. To combine the results from the static, dynamic, and sky fields, we sort samples along the ray based on their distance and rely on nerfacc~\cite{li2023nerfacc} for efficient rendering.

\subsection{Model components}
\label{sec:unisim:model}
\parsection{CNN}
The CNN used for upsampling consists of four residual blocks with 32 channels. Further, a convolutional layer is applied at the beginning of the CNN to convert input features to 32 channels, and a second convolutional layer is applied to predict the RGB values. For both layers, we use kernel size one with no padding. We set the residual blocks to consist of convolution, batch norm, ReLU, convolution, batch norm, and skip connection to the input. The convolutional layers in the residual block use a kernel size of seven, with a padding of three. Between the second and third residual blocks, we apply a transposed convolution to upsample the feature map. We set the kernel size and stride to the upsample factor. The upsample factor in turn is set to three. Although we follow the specifications of UniSim, we find our implementation to have fewer parameters than what they report (0.7M compared to 1.7M). Likely reasons are interpretations of residual block design (only kernel size and padding a specified), kernel size for the first and last convolution layer, and the design of the upsampling layer. Nonetheless, we found that increasing the CNN parameter count only increased run-time without performance gains.

\parsection{GAN}
UniSim deploys an adversarial training scheme, where a CNN discriminator is trained to distinguish between rendered image patches at observed and unobserved viewpoints, where unobserved viewpoints refer to jittering the camera origins. The neural feature field and upsampling CNN are then trained to improve the photorealism at unobserved viewpoints. UniSim results show adversarial training to hurt novel view synthesis metrics (PSNR, LPIPS, SSIM), but boost FID performance for the lane-shift setting.

Unfortunately, the discriminator design is only briefly described in terms of a parameter count, resulting in a large potential design space. As training is done on patches, we opted for a PatchGAN~\cite{isola2017image} discriminator design inspired by pix2pixhd~\cite{wang2018high}. However, we found it difficult to get consistent performance increases and hence removed the adversarial training from our reimplementation. This is likely the reason for our reimplementation to perform slightly worse than the original results in terms of FID for lane shift. However, using adversarial training does not seem to be necessary in general for achieving low FID scores. In \cref{tab:extrapolation_image}, we see NeuRAD, which does not use any GAN training, outperforming the original UniSim method, which does rely on adversarial supervision.

\parsection{SDF to occupancy mapping}
UniSim approximates the mapping from signed distance $s$ to occupancy $\alpha$ as
\begin{equation}
    \alpha = \frac{1}{1+e^{\beta s}},
\end{equation}
where $\beta$ is a hyperparameter. As $\beta$ is unspecified, we follow~\cite{yang2023reconstructing}, which uses a similar formulation for neural rendering in an automotive setting. Specifically, we initialize $\beta$ to 20.0 and make it a learnable parameter to avoid sensitivity to its specific value.

\subsection{Supervision}
\parsection{Loss hyperparameters}
We set $\lambda_\text{rgb}=1.0$ and $\lambda_\text{vgg}=0.05$. All other learning weights are given in UniSim's supplementary material and hence are used as is.

\parsection{Regularization loss}
For lidar rays, UniSim uses two regularizing losses. The first decreases the weights for samples far from any surface and the second encourages the signed distance function to satisfy the eikonal equation close to any surface
\begin{multline}
    \mathcal{L}_\text{reg} = \frac{1}{N}\sum_{i=1}^N \left(\sum_{\gamma_{i,j}>\epsilon}||w_{ij}||_2 \right. \\ \left. + \sum_{\gamma_{i,j}<\epsilon}(||\nabla s(\mathbf{x}_{ij})|| - 1)^2 \right).
\end{multline}
Here, $i$ is the ray index, $j$ is the index for a sample $\mathbf{x}_{ij}$ along the ray, $\gamma_{i,j}$ denotes the distance between the sample and the corresponding lidar observation, \ie, $\gamma_{i,j}=|\tau_{ij}-D_i^\text{gt}|$. We set $\epsilon=0.1$.

Furthermore, we rely on tiny-cuda-nn~\cite{tiny-cuda-nn} for fast implementations of the hash grid and MLPs. However, the framework does not support second-order derivatives for MLPs, and hence cannot be used when backpropagating through the SDF gradient $\nabla s(\mathbf{x}_{ij})$. Hence, instead of analytical gradients, we use numerical ones. Let
\begin{equation}
    \begin{bmatrix}
        \mathbf{k}_1 \\ \mathbf{k}_2 \\ \mathbf{k}_3 \\ \mathbf{k}_4
    \end{bmatrix}
    =
    \begin{bmatrix}
        1  & -1 & -1 \\
        -1 & -1 & 1  \\
        -1 & 1  & -1 \\
        1  & 1  & 1
    \end{bmatrix}.
\end{equation}
To find $\nabla s(\mathbf{x}_{ij})$, we query the neural feature field at four locations $\mathbf{x}_{ij}+\delta\mathbf{k}_l, l = 1,2,3,4$ where $\delta=\frac{0.01}{\sqrt{3}}$, resulting in four signed distance values $s_1, s_2, s_3, s_4$. Finally, we calculate
\begin{equation}
    \nabla s(\mathbf{x}_{ij}) = \frac{1}{4\delta}\sum_l s_l \mathbf{k}_l.
\end{equation}
The use of numerical gradients instead of analytical ones has been shown to be beneficial for learning signed distance functions for neural rendering~\cite{li2023neuralangelo}.

\parsection{Perceptual loss}
Just like NeuRAD, UniSim uses a perceptual loss where VGG features of a ground truth image patch are compared to a rendered patch. While multiple formulations of such a loss exist, we opted for the one used in pix2pixHD~\cite{wang2018high} for both methods.

\begin{figure}[th]
    \centering
    \begin{subfigure}{\linewidth}
        \includegraphics[trim={6.5cm 3.5cm 5cm 4cm},clip, width=1\linewidth]{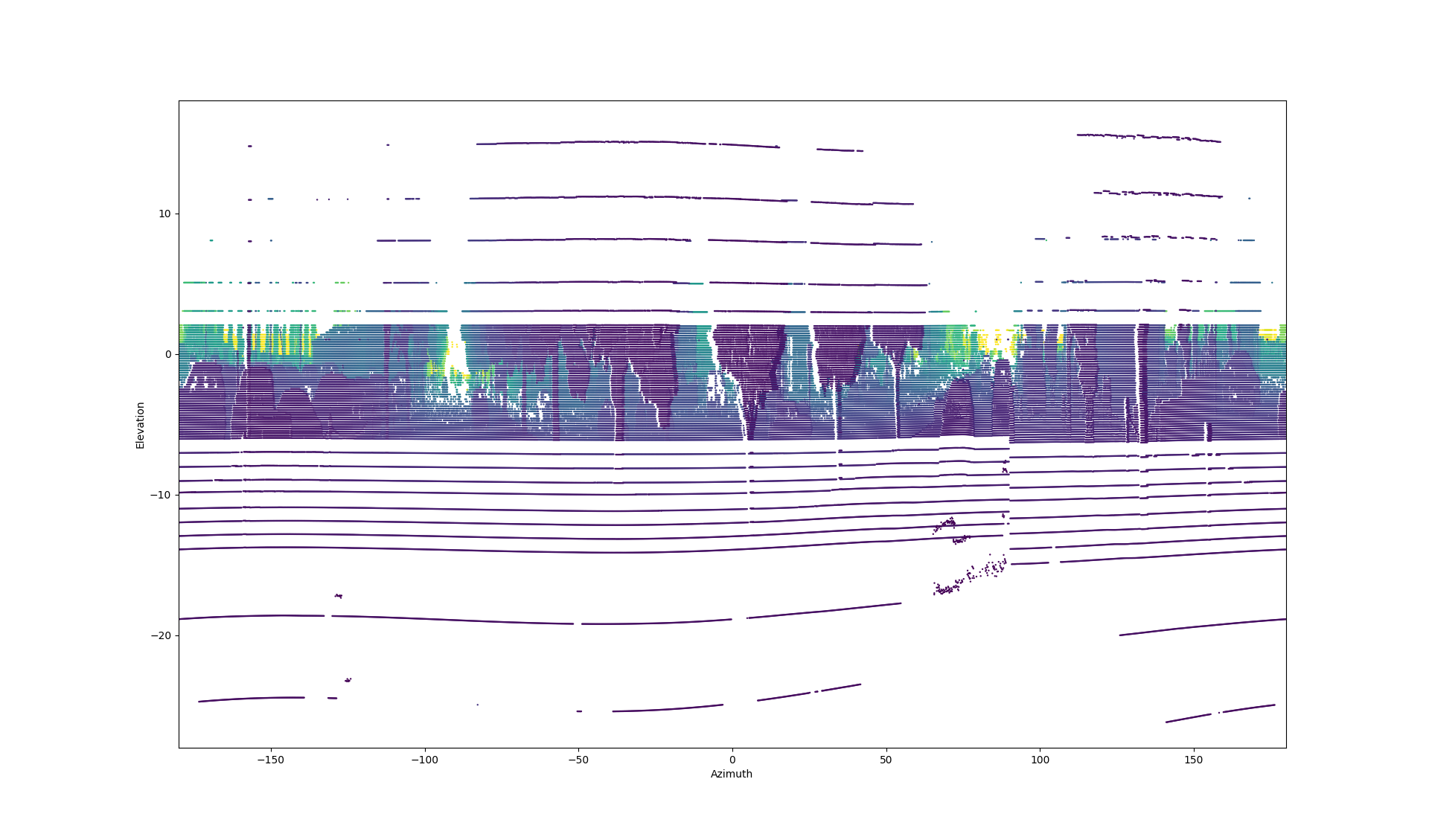}
        \caption{Before removing ego-motion compensation.}
    \end{subfigure}
    \begin{subfigure}{\linewidth}
        \includegraphics[trim={6.5cm 3.5cm 5cm 4cm},clip, width=1\linewidth]{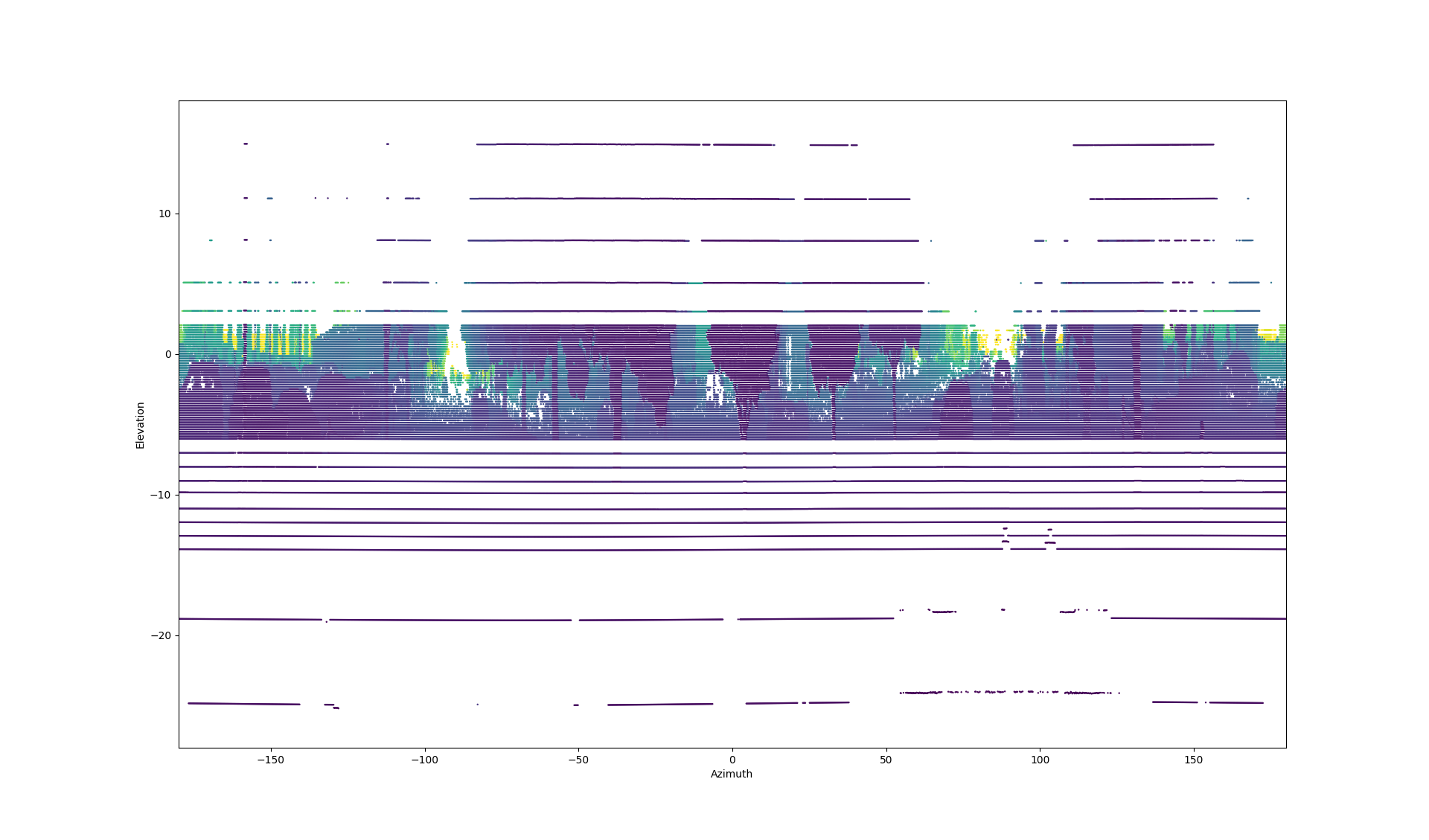}
        \caption{After removing ego-motion compensation.}
    \end{subfigure}
    \begin{subfigure}{\linewidth}
        \includegraphics[trim={6.5cm 3.5cm 5cm 4cm},clip, width=1\linewidth]{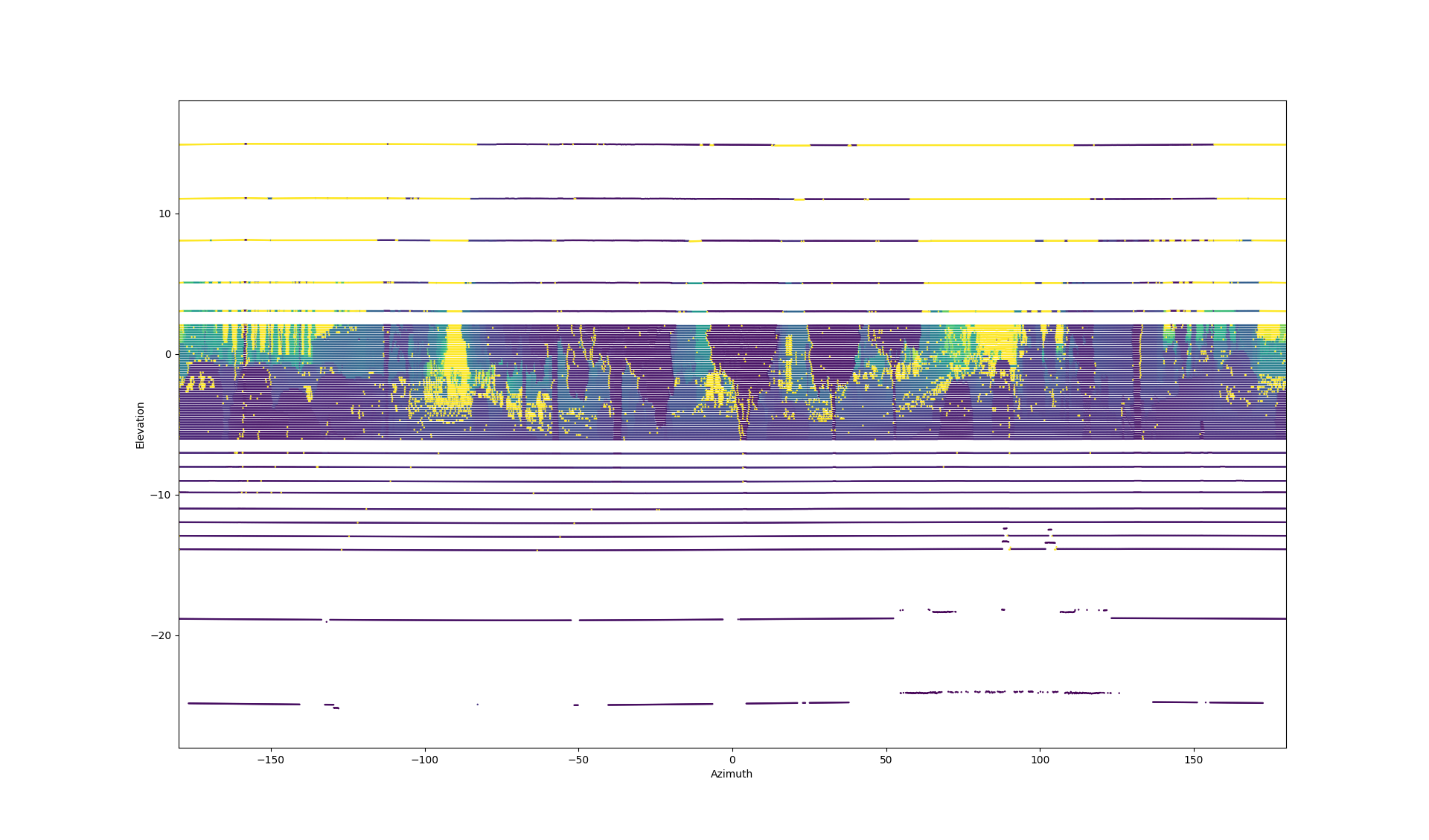}
        \caption{After removing ego-motion compensation and adding missing points.}
    \end{subfigure}
    \caption{Lidar scans in spherical coordinates at different stages during inference of missing lidar rays. The color indicates range, where missing points have been set to a large distance for visualization purposes. Note that we do not add missing points for the two bottom rows, as they typically hit the ego vehicle.}
    \label{fig:missing_points}
\end{figure}

\section{Inferring ray drop}
\label{sec:app_raydrop}

The inclusion of dropped lidar rays during supervision increases the fidelity of sensor renderings in all aspects, as shown in \cref{tab:ablations}. The process of inferring which lidar beams are missing in a point cloud differs somewhat between datasets, as they contain different types of information. However, in general, the process consists of three steps: removal of ego-motion compensation, diode index assignment, and point infilling. In \cref{fig:missing_points}, we show a lidar scan from PandaSet~\cite{pandaset} (sequence \texttt{106}) at different stages.

\parsection{Removal of ego-motion compensation} To figure out which points are missing in a single sweep, we want to express their location in terms of azimuth (horizontal angle), elevation (vertical angle), and range at the time the beam was shot from the sensor. However, for all datasets, the provided points have been ego-motion compensated, \ie, their Cartesian coordinates are expressed in a common coordinate frame. Simply converting them to spherical coordinates is therefore not possible until the ego-motion compensation is removed.

For each 3D lidar point $(x,y,z)$ captured at time $t$ we first project the point into world coordinates using its assigned sensor pose. For PandaSet~\cite{pandaset}, this first step is omitted, as points are provided in world coordinates. We then find the pose of the lidar sensor at time $t$ by linearly interpolating existing sensor poses. For rotation, we use a quaternion representation and spherical linear interpolation (slerp). Given the sensor pose at $t$, we project the 3D point back into the sensor frame. We note that this process is susceptible to noise, since lidar poses are typically provided at a low frequency \SI{10}{\hertz}-\SI{20}{\hertz}. We find elevation $\phi$, azimuth $\theta$ and range $r$ as
\begin{align}
    r      & = \sqrt{x^2+y^2+z^2},             \\
    \phi   & = \arcsin{\left( z / r \right)},  \\
    \theta & = \arctan{\left( y / x  \right)}.
\end{align}

\parsection{Diode index assigment}
All datasets considered in this work use spinning lidars, where a set of diodes are rotated 360$^\circ$ around the sensor and each diode is mounted at a fixed elevation angle. Typically, all diodes (or channels) transmit the same number of beams each revolution, where the number depends on the sensors' horizontal resolution. To use this information for inferring missing rays, we need to assign each return to its diode index. For most datasets considered here~\cite{caesar2020nuscenes,alibeigi2023zenseact,Argoverse2}, this information is present in the raw data. However, for the other~\cite{pandaset}, we must predict diode assignment based on the points' elevation. As there is no ground truth available for this information, we use qualitative inspections to verify the correctness of the procedures outlined below.

PandaSet uses a spinning lidar with a non-linear elevation distribution for the diodes, \ie, diodes are not spaced equally along the elevation axis. Instead, a few channels, the ones with the largest and smallest elevations, have a longer distance from their closest diode neighbor. Points corresponding to these channels are easily found by using sensor specifications. The remaining diodes use equal spacing, but inaccuracies in the removal of ego-motion compensation result in many wrongful diode assignments if sensor specifications are trusted blindly. Thus, we devise a clustering algorithm for inferring diode indices for points originating from diodes within the equal elevation distribution range.

The following is done separately for each lidar scan. First, we define the expected upper and lower bounds for elevation for each diode. These decision boundaries are spaced equally between the lowest and highest observed elevations based on the number of diodes. Then, we use histogram binning to cluster points based on their elevation. We use 2,000 bins, and the resulting bin widths are smaller than the spacing between diodes. Next, we identify consecutive empty bins. For any \textit{expected} decision boundary that falls into an empty bin, we mark it as a \textit{true} decision boundary. The same is true if the expected decision boundary is within $0.03^\circ$ of an empty bin. Following this, if the number of true decision boundaries is smaller than the number of expected decision boundaries, we insert new boundaries between existing ones. Specifically, for the two boundaries with the largest distance between them, we insert as many boundaries as the vertical resolution dictates, but at least one, and at most as many decision boundaries that are missing. This insertion of boundaries is repeated until the required number of boundaries is reached.

\parsection{Point infilling}
After removing ego-motion compensation, transforming the points to spherical coordinates (elevation, azimuth, range), and finding their diode index, we can infer which laser rays did not return any points. Separately, for each diode, we define azimuth bins, spanning $0^\circ$ to $360^\circ$ with a bin width equal to the horizontal resolution of the lidar. If a returning point falls into a bin, we mark it as returned. For the remaining bins, we calculate their azimuth and elevation by linear interpolation.

\section{Modeling rolling shutter}
\label{sec:rolling_shutter}
\begin{figure}[t]
    \centering
    \includegraphics[width=\linewidth]{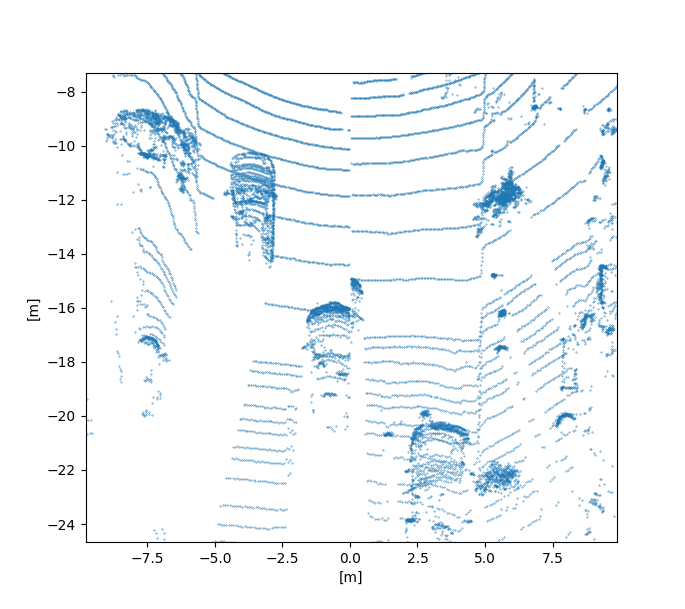}
    \caption{Bird's-eye-view of ego-motion compensated point cloud. Cuts in the circular patterns on the ground indicate the distance traveled by the ego-vehicle during one lidar revolution. Further, the cut through the car shows the importance of interpolating actor poses to the time when each lidar ray was shot.}
    \label{fig:lidar_rollingshutter}
\end{figure}

As shown in \cref{fig:qualitative results} and \cref{tab:ablations}, modeling the rolling shutter improves generated renderings, especially at high velocities. \cref{fig:lidar_rollingshutter} further shows the effects of rolling shutter on an ego-motion compensated lidar point cloud. To capture these effects, we assign each ray an individual timestamp. For lidar, these timestamps are typically available in the raw data, else we approximate them based on the rays' azimuth and the sensors' RPM. For cameras, individual timestamps are not available in the data. Instead, we manually approximate the shutter time and offset each image row accordingly. Given the individual timestamps, we linearly interpolate sensor poses to these times, effectively shifting the origin of the rays. Moreover, we model that dynamic actors may move during the capture time. Given the timestamps, we linearly interpolate their poses to the said time before transforming ray samples to the actors' coordinate systems.

\begin{figure*}
    \centering
    \includegraphics[width=\linewidth]{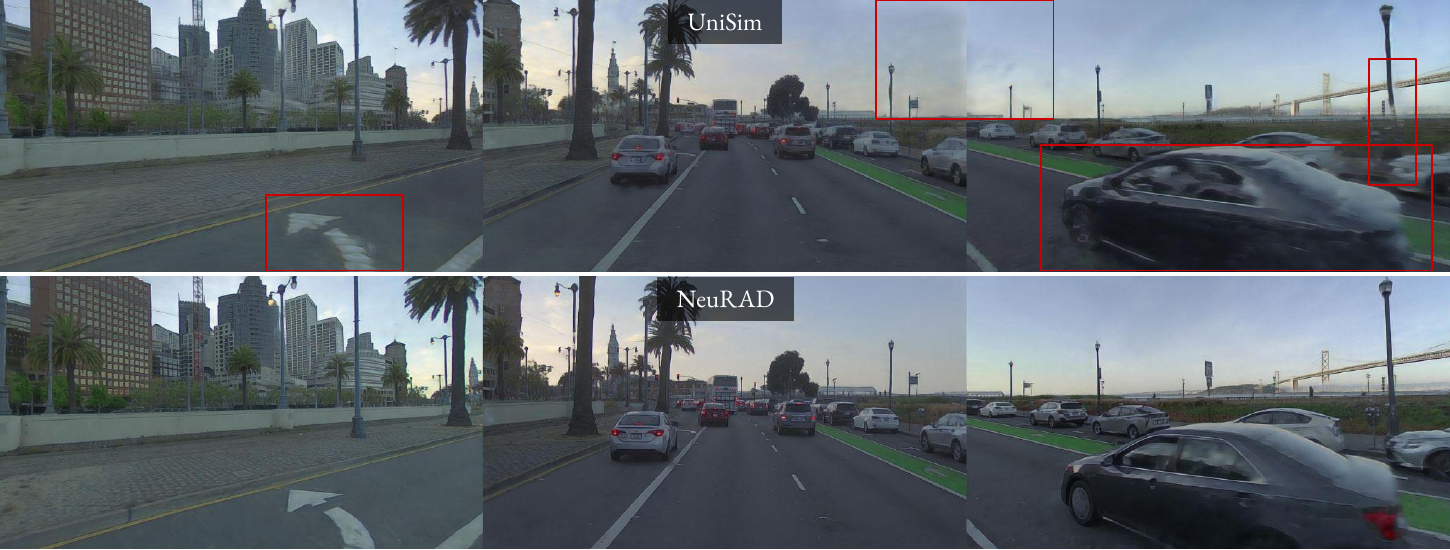}
    \includegraphics[width=\linewidth]{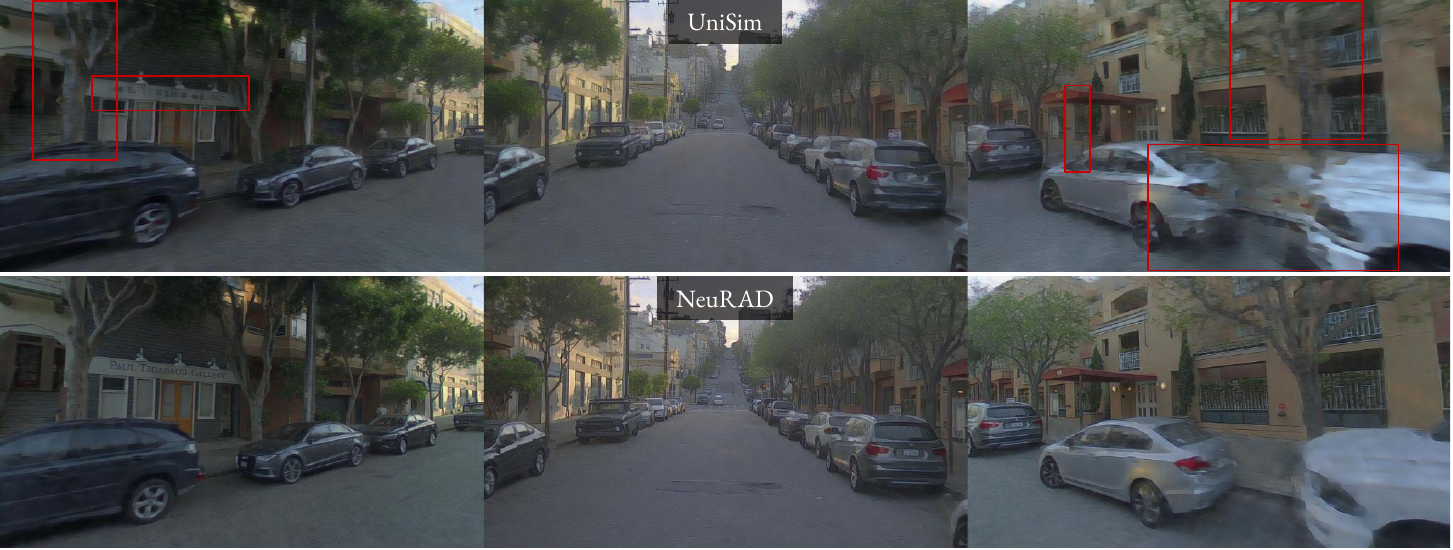}
    \includegraphics[width=\linewidth]{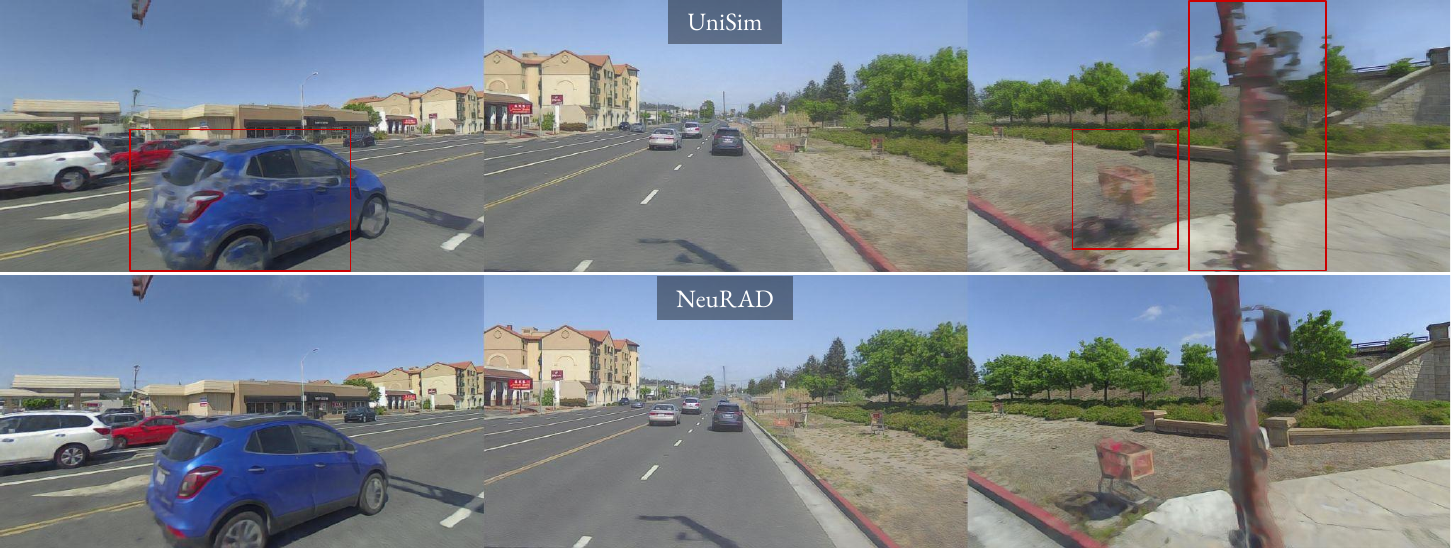}
    \caption{Qualitative comparison between NeuRAD and UniSim across three Pandaset sequences (016, 028, 158). Displayed are the front left, front center, and front right camera perspectives. NeuRAD overall captures more details than UniSim, although the difference is not dramatic for the front camera. However, as highlighted by red boxes, NeuRAD clearly outperforms UniSim for side cameras.}
    \label{fig:vs_unisim}
\end{figure*}

\section{Simulation gap}
\label{sec:simulation_gap}
In the following, we show results for the simulation gap. To study the real2sim gap, we train the 3D object detector BEVFormer on real images from PandaSet-360 and evaluate its object detection performance on synthesized images from the ten sequences used for NVS (not part of the training set for BEVFormer). For BEVFormer, we use the official implementation\footnote{\url{https://github.com/fundamentalvision/BEVFormer}} and the small version of the model. In \cref{tab:real2sim} we see that the detector achieves similar validation performance for the real images and the synthesized images from NeuRAD. UniSim, struggling in the 360 setting, exhibits a larger gap.

Further, for a more general and dataset-agnostic evaluation, we use a zero-shot depth estimator, DepthAnything (DA). We measure the agreement between depth estimations on synthesized and real images using the standard $\delta_1$ metric. \cref{tab:model_performance} shows consistent depth estimations, indicating a low domain gap across several datasets. For reference, DA reports $\delta_1\!=\!0.947$ when comparing against ground truth depth. We find these studies to give valuable insights and will include them in the manuscript.
\begin{table}[t]
    \centering
    \caption{Real2Sim gap: BEVFormer (mAP) on different images.}
    \vspace{-2mm}
    \resizebox{0.7\linewidth}{!}{%
    \begin{tabular}{c|cc|c}
    Data source &  NeuRAD & UniSim$^*$ & Real \\ \hline
    mAP & \textbf{32.0}   & 30.1 & 32.4 
    \end{tabular}
    }
    \vspace{-0mm}
    \label{tab:real2sim}
\end{table}

\begin{table}[t]
\centering
\caption{Real2Sim gap: DepthAnything relative depth ($\delta_1\!\uparrow$).}
\vspace{-2mm}
\setlength{\tabcolsep}{4pt} 
\resizebox{\linewidth}{!}{%
\begin{tabular}{l|cccccc}
        & PandaFC & Panda360 & AV2   & ZOD    & NuScenes & KITTI     \\ \hline
UniSim$^*$  & 0.927    & 0.872     & 0.860 & 0.901  & -        & -         \\
Neurad  & \textbf{0.968}    & \textbf{0.944}     & \textbf{0.928} & \textbf{0.958}  & \textbf{0.894}    & \textbf{0.947}
\end{tabular}
}
\vspace{-0mm}
\label{tab:model_performance}
\end{table}

\section{Additional results}
\label{sec:more_results}
In the following, we provide additional results and insights, as well as some failure cases of our method.

\parsection{Comparison with UniSim}
We begin with a direct qualitative comparison between NeuRAD and UniSim~\cite{unisim} as depicted in \cref{fig:vs_unisim}. For the front camera, the distinction in quality is subtle but observable; NeuRAD demonstrates superior image clarity, exhibiting notably less noise and artifact presence. In contrast, the disparity in quality is more pronounced with the side cameras. Here, NeuRAD's output markedly surpasses UniSim, which is particularly evident in the highlighted areas where UniSim exhibits significant motion blur and visual distortion that NeuRAD effectively mitigates.

\begin{figure*}[t]
    \centering
    \includegraphics[width=\linewidth]{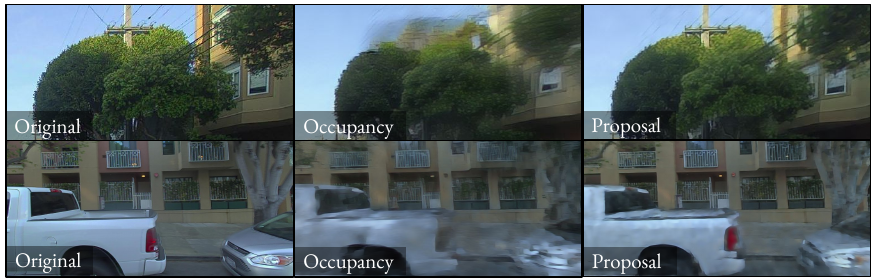}
    \caption{Two failure-cases that demonstrate the importance of proposal sampling over occupancy-based sampling: regions without lidar occupancy that are improperly modeled by sky field (upper), and nearby object that require extremely dense sampling (lower). }
    \label{fig:proposal_sampling}
\end{figure*}

\parsection{Proposal sampling}
To efficiently allocate samples along each ray, we use two rounds of proposal sampling. For comparison, UniSim instead samples along the rays uniformly and relies on a lidar-based occupancy grid to prune samples far from the detected surfaces. Although the occupancy grid is fast to evaluate, it has two shortcomings. First, the method struggles with surfaces far from any lidar points. In the case of UniSim, the RGB values must instead be captured by the sky field, effectively placing the geometry far away regardless of its true position. The upper row of \cref{fig:proposal_sampling} shows an example of this, where a utility pole becomes very blurry without proposal sampling. Second, uniform sampling is not well suited for recovering thin structures or fine details of close-up surfaces. Doing so would require drawing samples very densely, which, instead, scales poorly with computational requirements. We examine both failure cases in \cref{fig:proposal_sampling}, with thin power lines in the upper row and close-ups of vehicles in the lower row.

\begin{figure*}[t]
    \centering
    \includegraphics[width=\linewidth]{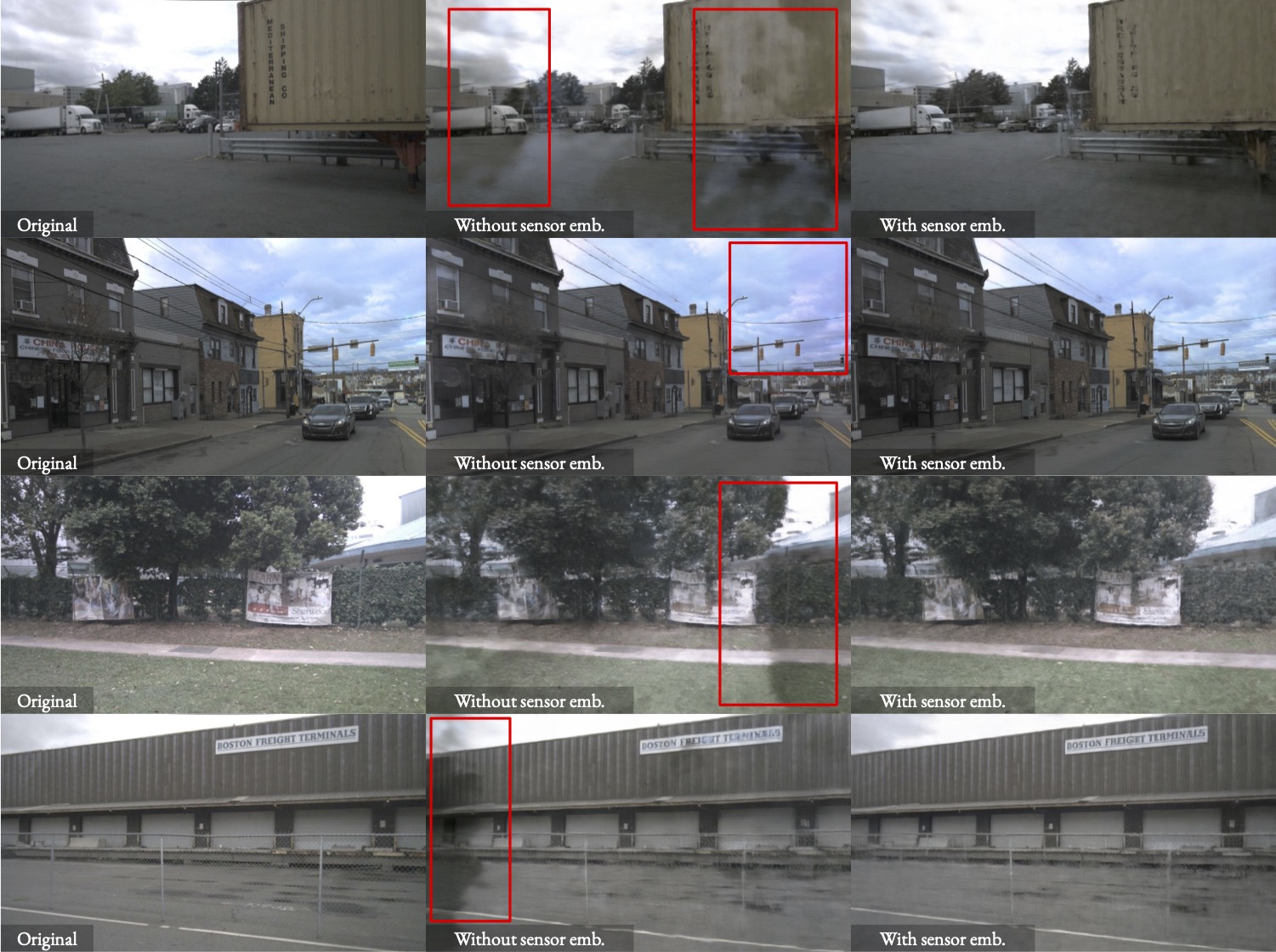}
    \caption{Effect of sensor embedding. The second column shows rendered images from the model trained without sensor embeddings, where a clear degradation is visible due to the shift in appearance (e.g., different exposure) between different sensors. As can be seen in the third column, this effect is remedied by including sensor embeddings.}
    \label{fig:appemb}
\end{figure*}

\parsection{Sensor embedding}
As described in \cref{sec:auto_data_modeling}, and shown in \cref{tab:ablations}, the effect of different camera settings for different sensors in the same scene has a significant impact on reconstruction results. \cref{fig:appemb} shows qualitative results of this effect. Ignoring this effect causes shifts in color and lighting, often at the edge of images where the overlap between sensors is bigger, and is clearly visible in the second column of \cref{fig:appemb}. Including sensor embeddings allows the model to account for differences in the sensors (e.g., different exposure), resulting in more accurate reconstructions.

\parsection{Camera optimization}
Neural rendering is reliant on access to accurate sensor poses. For instance, a small translation or rotation of a camera in world coordinates might translate to a small shift in the image plane as well, but this can drastically change each pixels' value.

In this work, we rely on sensor poses provided in the datasets, which typically are the result of IMU and GPS sensor fusion, SLAM, or a combination of both. As a result, sensor poses are often accurate to centimeter precision. While nuScenes~\cite{caesar2020nuscenes} follows this example, the dataset does not provide height, roll, or pitch information, as this information has been discarded. We found this to be a limiting factor for the performance of NeuRAD, especially for sequences where the ego vehicle does not traverse a simple, flat surface. To address this, we instead enable optimization of the sensor poses, similar to how we optimize the poses of dynamic actors, see \cref{sec:auto_data_modeling}.

\begin{figure*}[t]
    \centering
    \includegraphics[width=\linewidth]{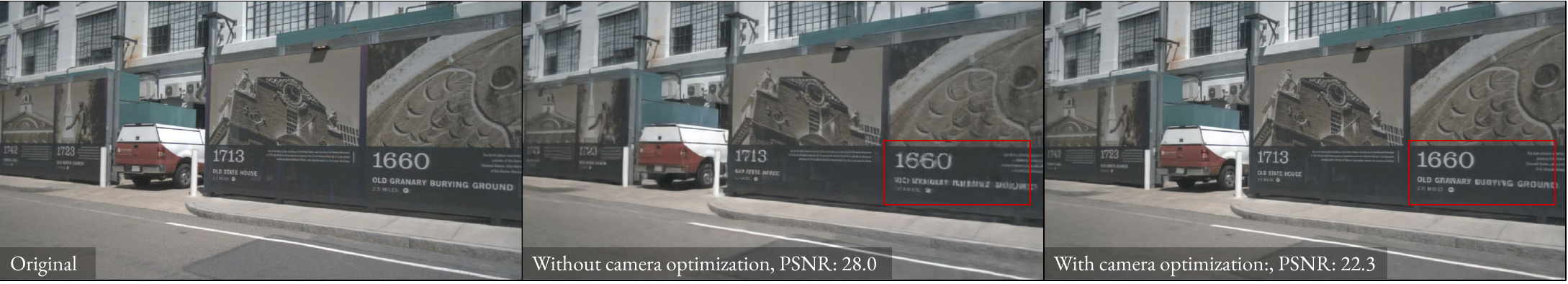}
    \caption{Effect of camera optimization on nuScenes. Despite clearly sharper image quality, we get drastically reduced PSNR scores when using camera optimization. This is due to the misalignment between the learned poses and the evaluation poses. This can be seen in the far left of the image, where the image with camera optimization displays less of a window.}
    \label{fig:camopt2}
\end{figure*}

Applying sensor pose optimization qualitatively results in sharp renderings and quantitatively yields strong FID scores, see \cref{tab:extrapolation_image}. However, we found novel view synthesis performance -- in terms of PSNR, LPIPS and SSIM -- to drop sharply. We find that the reason is that the sensor pose optimization creates an inconsistency between the world frame of the training data and the validation poses. Due to noisy validation poses, we render the world from a slightly incorrect position, resulting in large errors for the NVS per-pixel metrics. We illustrate this in \cref{fig:camopt2}, where the image from the training without sensor pose optimization is more blurry, but receives higher PSNR scores than the one with pose optimization.

We explored multiple methods for circumventing these issues, including separate training runs for finding accurate training and validation poses, or optimizing only the poses of validation images post-training. However, to avoid giving NeuRAD an unfair advantage over prior work, we simply disabled sensor pose optimization for our method. Nonetheless, we hope to study the issue of NVS evaluation when accurate poses are not available for neither training or validation in future work.

\begin{figure*}[t]
    \centering
    \includegraphics[width=\linewidth]{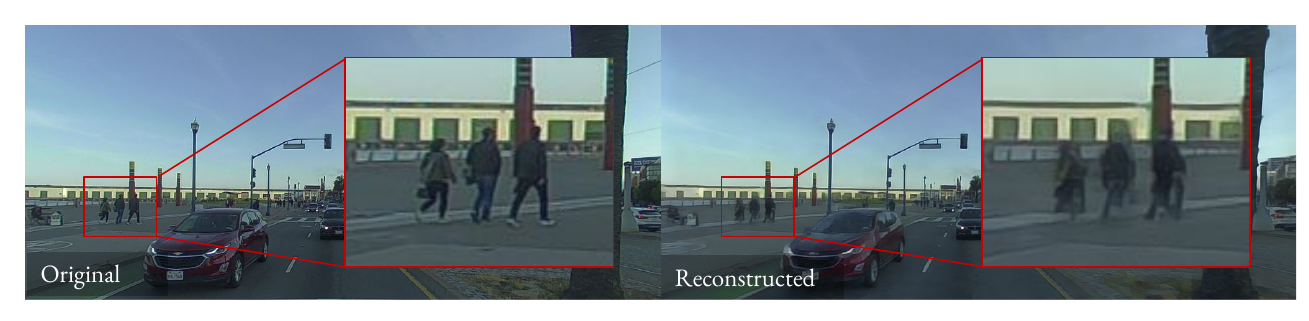}
    \caption{Failed reconstruction of deformable actors. The assumption that all actors are rigid is invalid for pedestrians and the like, leading to blurry reconstruction as seen here.}
    \label{fig:pedestrians}
\end{figure*}

\subsection{Limitations}
In this work, we have proposed multiple modeling strategies for capturing important phenomena present in automotive data. Nonetheless, NeuRAD builds upon a set of assumptions, which when violated, result in suboptimal performance. Here, we cover some of these failure cases.

\parsection{Deformable dynamic actors}
When modeling dynamic actors, we make one very strong assumption --- that the dynamics of an actor can be described by a single rigid transform. This is a reasonable approximation for many types of actors, such as cars, trucks, and to a lesser degree even cyclists. However, pedestrians break this assumption entirely, leading to very blurry reconstructions, as can be seen in \cref{fig:pedestrians}.

\parsection{Night scenes}
Modelling night scenes with NeRF-like methods can be quite tricky for several reasons. First, night images contain a lot more measurement noise, which hinders the optimization as it is not really related to the underlying geometry. Second, long exposure times, coupled with the motion of both the sensor and other actors, lead to blurriness and can even make thin objects appear transparent. Third, strong lights produce blooming and lens-flare, which have to be explained by placing large blobs of density where there should not be any. Finally, dynamic actors, including the ego-vehicle, frequently produce their own illumination, such as from headlights. While static illumination can usually be explained as an effect dependent on the viewing direction, this kind of time-varying illumination is not modelled at all.

\begin{figure*}[t]
    \centering
    \includegraphics[width=\linewidth]{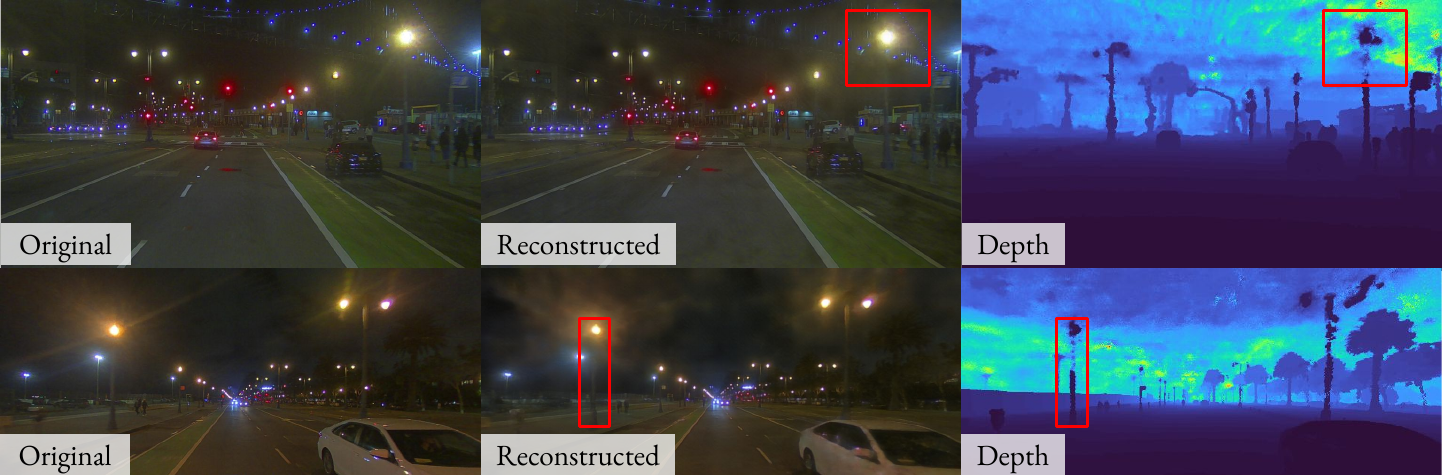}
    \caption{Novel view synthesis at night is challenging. For instance, strong lights can produce flares in the camera lens. These are hard to model with the standard NeRF rendering equations, as it requires the network to place density around the lights. Further, longer exposure times at night lead to dark, thin objects appearing semi-opaque, obscuring the learned scene geometry. Last, moving vehicles, including the ego-vehicle, illuminate the scene, resulting in a change of color over time for certain static parts of the scene. For instance, the road contains artifacts due to illumination from the ego-vehicle headlights.}
    \label{fig:at-night}
\end{figure*}

\parsection{Time-dependent object appearance}
In order to build a fully-useable closed-loop simulation we need to model brake lights, turning indicators, traffic lights, etc. While the problem is similar to that of deformable actors, it differs in some ways. First, we do not require the geometry to vary over time, potentially simplifying the problem. Second, we can probably treat these appearances as a set of discrete states. Third, the current set of perception annotations/detections might not cover all necessary regions where this effect is present. For instance, most datasets do not explicitly annotate traffic lights. Finally, we require full control and editability for this effect, to the degree that we can enable brake lights for a car that never braked. For general deformable actors, we might be satisfied with reconstructing the observed deformation, without being able to significantly modify it.

\begin{figure*}[t]
    \centering
    \includegraphics[width=\linewidth]{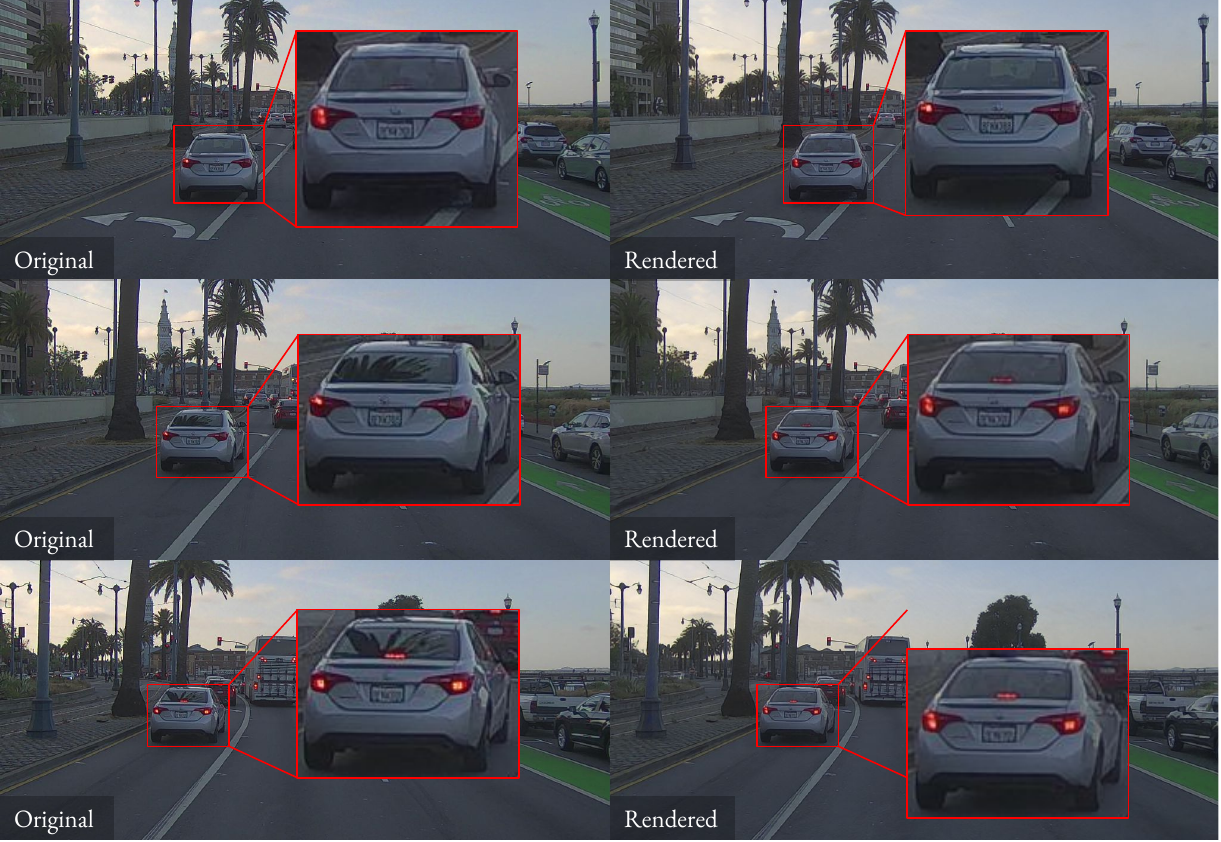}
    \caption{NeuRAD assumes all radiance to be static over time, even for dynamic actors. Thus, our method cannot express changes in light conditions, such as brake lights highlighted here. Interestingly, the model compensates by making the brake lights a function of the viewing angle instead, as the two are correlated in this particular scene.}
    \label{fig:brakelight}
\end{figure*}

\end{document}